\PassOptionsToPackage{svgnames}{xcolor}   
\documentclass[10pt,twocolumn,letterpaper]{article}
\usepackage[preprint]{cvpr}      
\usepackage{graphicx}
\usepackage{amsmath,amssymb}
\usepackage{booktabs}
\usepackage{caption}
\usepackage{adjustbox}   
\graphicspath{{figs/}{./}}
\usepackage{booktabs}
\usepackage{multirow}
\usepackage{amsmath,amssymb}
\usepackage{bm}
\usepackage[svgnames]{xcolor} 
\usepackage{colortbl}
\definecolor{oursbg}{HTML}{FAD2E4}
\definecolor{oursfg}{HTML}{D9468C}
\newcommand{\oursrow}{\rowcolor{oursbg}}
\usepackage{algorithm}
\usepackage{algpseudocode}
\usepackage{pifont}

\newcommand{\ours}{AdaPixFlow}

\DeclareRobustCommand{\appref}[1]{\iflabelexists{#1}{\cref{#1}}{the appendix of the full paper in the supplemental material}}

\newif\ifdraftmarkers \draftmarkerstrue
\ifdraftmarkers
  \newcommand{\todo}[1]{\textcolor{red}{[TODO: #1]}}
  
\else
  
  \newcommand{\todo}[1]{}
  
\fi

\newcommand{\up}{$\uparrow$}
\newcommand{\down}{$\downarrow$}
\newcommand{\best}[1]{\textbf{#1}}
\newcommand{\second}[1]{\underline{#1}}

\newcommand{\z}{\bm{z}}
\newcommand{\X}{\mathcal{X}}
\newcommand{\V}{\bm{V}}
\newcommand{\vel}{\bm{v}}
\newcommand{\ctx}{\bm{c}}
\newcommand{\R}{\mathbb{R}}
\newcommand{\Lift}{\mathcal{L}}

\usepackage{placeins} 
\makeatletter\providecommand{\iflabelexists}[3]{\@ifundefined{r@#1}{#3}{#2}}\makeatother   
\definecolor{cvprblue}{rgb}{0.21,0.49,0.74}
\usepackage[pagebackref,breaklinks,colorlinks,allcolors=cvprblue]{hyperref}
\usepackage[capitalize,noabbrev]{cleveref}
\AtBeginDocument{%
  \Crefname{appendix}{Appendix}{Appendices}%
  \Crefname{subappendix}{Appendix}{Appendices}%
  \Crefname{subsubappendix}{Appendix}{Appendices}%
}

\def\confName{CVPR}
\def\confYear{2027}
\title{\ours: Learning Continuous-Time 3D Dynamics Across Scenes from Sparse Multi-view Observations}
\author{Hrishikesh Viswanath\qquad Aniket Bera\\
Department of Computer Science, Purdue University\\
{\tt\small hviswan@purdue.edu}\\
{\small\url{https://github.com/HrishikeshVish/adapixflow}}}

\begin{document}
\twocolumn[{%
\renewcommand\twocolumn[1][]{#1}%
\maketitle
\begin{center}
  \centering
  \includegraphics[width=\linewidth]{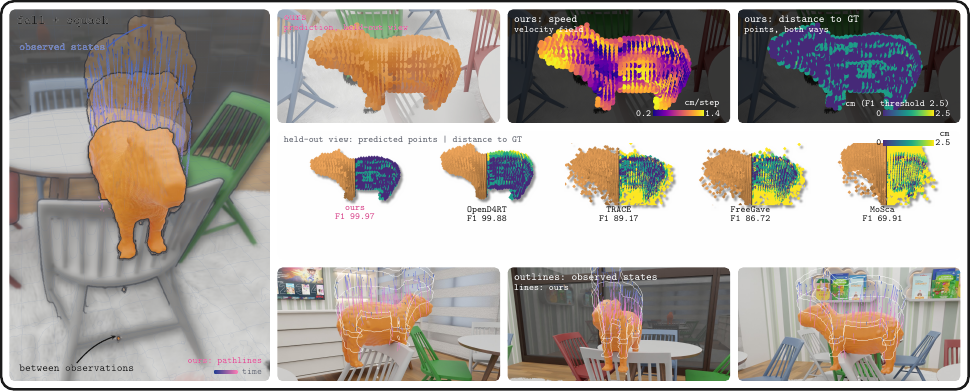}
  \captionof{figure}{\ours{} on a held-out scene, PlasticineFall. Left: a sketch of observed states and our pathlines. Right: a held-out view (top and middle rows) and three observed views (bottom row) at a step between observations. Middle row: predicted points of \ours{}, OpenD4RT, TRACE, FreeGave, and MoSca, shaded by distance to the ground truth (darker is closer).}
  \label{fig:teaser}
\end{center}
\vspace{1em}
}]

\begin{abstract}
Understanding how objects move in dynamic scenes requires recovering their 3D motion from visual observations and predicting their positions at unobserved times. Existing methods either optimize scene-specific velocity fields or impose predefined physical and kinematic assumptions. We instead study learning a continuous-time dynamics operator across scenes, mapping multi-view visual observations to 3D object kinematics. We introduce Adaptive Pixel Flows (\ours{}), which learns this operator from multi-view RGB-D observations without per-scene optimization or predefined motion models. The method fuses observations from all cameras into a shared voxel grid in world coordinates, from which the network predicts a continuous-time velocity field. Because the operator is trained across scenes, it learns motion priors that carry over to unseen scenes. The field can then be integrated from observed states to predict object positions between observations and beyond the observed sequence. On held-out PhysInOne scenes, \ours{} improves F1 over the strongest competing method by 2.5\% for in-between prediction and 15\% for future prediction. Compared with per-scene velocity-field methods, the improvements are 30\% and 85\%, respectively. Applied zero-shot to Spring-Gaus, it improves future-prediction F1 by 12\%.
\end{abstract}

\section{Introduction}
\label{sec:intro}
Predicting how objects move, such as a swinging pendulum, a spinning panel, or an object dropping into liquid, is a key task in animation, simulation, mixed reality, and robotics \cite{levine2012physically, chen2026agentic, agrawal2016learning, dou2026neuralactuator, lai2026diffphd,xiong2025topogaussian, ma2023learning}. Dynamic reconstruction methods recover such motion from video by fitting a time-varying 3D representation to each scene. This can include deformed canonical space, factorized spacetime volume, or trajectories of 3D Gaussians~\cite{pumarola2021dnerf,park2021nerfies,cao2023hexplane,fridovichkeil2023kplanes,yang2024deformable,wu20244d,luiten2024dynamic,wang2025shape,lei2025mosca}. Feed-forward 3D trackers~\cite{xiao2025spatialtrackerv2} and 4D reconstruction models~\cite{zhang2026efficiently} infer time-varying geometry and correspondences across scenes without per-scene fitting. These methods can be queried at any time within the observed sequence, but they do not model the dynamics that would predict how the scene evolves after it.

A complementary line of work focuses on explicitly modeling scene dynamics. Some methods represent motion through learned velocity fields~\cite{li2023nvfi,li2025freegave,lu2026velogauss} or per-particle dynamics~\cite{li2025trace}. Others recover material parameters from visual observations to drive differentiable physical simulators~\cite{li2023pacnerf,zhong2024reconstruction,jiang2025phystwin}, or specify these parameters manually~\cite{xie2024physgaussian}. While these approaches provide a more explicit account of temporal evolution, they are often optimized independently for each scene and rely on a prescribed model of the underlying dynamics. Recent video-based dynamics models~\cite{zhobro2026learning,zhang2025particle,wang2026neural} learn dynamics that transfer across scenes or object instances, but remain tied to specific material or object categories, or rely on pretrained world-model representations. 

A different modeling perspective is to treat 3D dynamics as a direct parameterization of visual observations, rather than scene-specific or material-specific models. In this paradigm, one can learn the representation space for a continuous-time dynamics operator that can generalize across physical systems and scenes. Inspired by the operator-learning philosophy \cite{li2021fourier, lu2021learning, viswanath2026neural}, we seek a representation of such an operator that admits zero-shot evaluation on previously unseen scenes, without material-specific modeling.

Concretely, we consider kinematics through a continuous-time 3D velocity field that describes the evolution of object positions. Learning such a field directly from video presents a fundamental challenge: the motion between and after observations may be underdetermined by the observations themselves, because visual observations may be entangled with viewpoint and occlusion or be sparsely sampled. Thus, the field must draw on priors learned across scenes. Addressing the entanglement with viewpoint and occlusion further requires a representation in which scene geometry and motion can be expressed independently of the viewing configuration.

Our key insight is to learn the dynamics operator on a volumetric fusion of the observations. This expresses scene states in one world coordinate frame shared by all cameras, rather than in the image coordinates of each view. We draw from the longstanding use of volumetric fusion to combine multiple views into a common representation, from depth fusion~\cite{curless1996volumetric,newcombe2011kinectfusion} to perception across arbitrary camera rigs~\cite{philion2020lift}. With calibrated multi-view RGB-D observations providing the geometric grounding, we describe scene evolution through the 3D motion of back-projected visual observations, which we term \emph{Adaptive Pixel Flows (AdaPixFlow)}. As illustrated in \cref{fig:pipeline}, AdaPixFlow parameterizes this motion using an observation-conditioned continuous-time 3D velocity field, along which the back-projected object points are evolved through differentiable advection. This enables continuous-time prediction of object positions both between observations and beyond the observed sequence. \Cref{fig:teaser} shows predictions on a held-out scene.

Our contributions are:
\begin{itemize}
    \item We introduce a continuous-time dynamics operator that, trained once across scenes, predicts the 3D motion of objects in unseen scenes from sparse multi-view RGB-D observations.
    \item We formulate the dynamics on a fused voxel grid, in which the back-projected object pixels are advected along a predicted velocity field, and train it through the advection without velocity labels or ground-truth point correspondences.
     \item We evaluate \ours{} on held-out PhysInOne~\cite{zhou2026physinone} scenes and test its zero-shot generalization on Spring-Gaus~\cite{zhong2024reconstruction}, PhysGaia~\cite{kim2026physgaia}, and Kubric-4D~\cite{vanhoorick2024gcd,greff2022kubric}. Across these benchmarks, we compare against per-scene dynamics models and adapted 4D reconstruction and tracking methods. On PhysInOne, \ours{} improves F1 over the strongest competing method by 2.5\% for in-between prediction and 15\% for future prediction, with gains of 30\% and 85\% over per-scene velocity-field methods. Without additional training, it improves future-prediction F1 on Spring-Gaus by 12\% and in-between prediction F1 on PhysGaia by 5\%, while achieving competitive performance on Kubric-4D, including against methods trained on that benchmark.
\end{itemize}
\section{Related Work}\label{sec:related}

\textbf{Dynamic reconstruction and tracking.} Neural radiance fields and 3D Gaussian splatting have enabled dynamic scene representations based on canonical-space deformations, factorized spacetime representations, and time-dependent Gaussian trajectories~\cite{pumarola2021dnerf,park2021hypernerf,fang2022fast,cao2023hexplane,yang2024realtime,luiten2024dynamic}. Subsequent work recovers 4D geometry and motion from monocular video~\cite{wang2025shape,lei2025mosca}, while feed-forward methods extend reconstruction and tracking to unseen scenes~\cite{xiao2025spatialtrackerv2,harley2025alltracker,zhang2026efficiently}. More recently, Trace Anything~\cite{liu2026trace} and Uni4R~\cite{yang2026marrying} have introduced continuous-time 3D point queries. Trace Anything predicts, for each pixel, a cubic B-spline trajectory with a fixed number of control points over the normalized time span of its input frames, which may be sparse or unordered. It forecasts by continuing a trajectory along its tangent rather than with a learned dynamics model. Uni4R predicts per-pixel 3D velocities from features interpolated between two consecutive frames and integrates them over that interval. Under this frame-pair formulation, every query needs the frame after it, so Uni4R does not predict beyond the last input frame. \ours{} instead learns a continuous-time dynamics operator in a shared 3D space, enabling motion prediction both between sparse observations and beyond the observed sequence.

\textbf{Learned dynamics models.} Another line of work learns predictive dynamics from visual observations using latent particles, RGB-D particle grids, or object slots~\cite{whitney2024learning,zhobro2026learning,zhang2025particle,wu2023slotformer}. Related approaches model temporal evolution through continuous-time neural dynamics~\cite{chen2018neural,park2021vidode} or learned operators~\cite{li2021fourier,chen2023crom}. Several of these models operate on explicit 3D particle or Gaussian states~\cite{whitney2024learning,zhobro2026learning,zhang2025particle} and typically advance them in discrete simulation steps. \ours{} combines a shared metric 3D state, fused from all calibrated views, with a continuous-time dynamics operator that can be queried at any time between and after observations.

\textbf{Physical dynamics from video.} Physical dynamics can also be recovered from video using physics-informed velocity fields~\cite{li2023nvfi} or differentiable simulators of continuum mechanics and spring-mass systems~\cite{li2023pacnerf,cai2024gic,zhong2024reconstruction,jiang2025phystwin}. Such methods either estimate simulator parameters from observations or take them as given~\cite{xie2024physgaussian}. Other approaches estimate per-Gaussian velocities~\cite{li2025freegave,li2025trace,lu2026velogauss}. These methods typically optimize scene-specific representations under prescribed dynamics formulations. In contrast, \ours{} learns a continuous-time dynamics operator across scenes and infers a 3D velocity field for a new scene directly from its observations, without scene-specific optimization or a predefined dynamics model.

\section{Method}\label{sec:method}
\begin{figure*}[t]
  \centering
  \includegraphics[width=\linewidth]{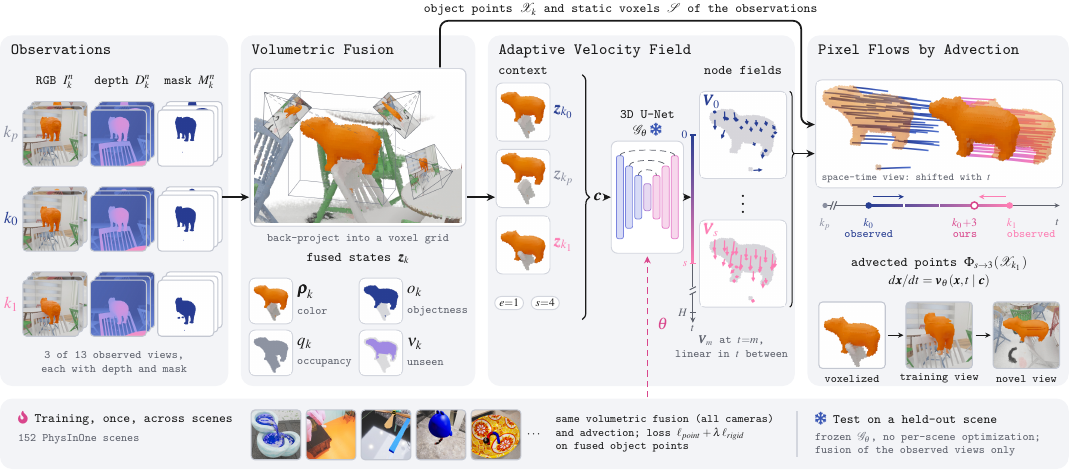}
  \caption{Overview of \ours{} on the held-out PlasticineFall scene (in-between prediction, $s{=}4$). Top: Multi-view observations are fused volumetrically and processed by the network $\mathcal{G}_\theta$ to predict node velocity fields $\V_0$ and $\V_s$, which are used to obtain pixel flows at $k_0{+}3$ through advection. Bottom: The operator is trained across scenes and evaluated on a held-out scene.}
  \label{fig:pipeline}
\end{figure*}

\Cref{sec:formulation} defines the problem and fuses the multi-view observations into a voxel grid. \Cref{sec:operator} defines the continuous-time dynamics operator, \cref{sec:advect} advects object points along its velocity field, and \cref{sec:training} describes training.

\subsection{Problem Formulation}\label{sec:formulation}

\textbf{Setup.} Let $\mathcal{T}=\{0,\ldots,T-1\}$ denote the uniformly spaced time steps of a clip, with observations available at $\mathcal{K}\subseteq\mathcal{T}$. The scene is captured by $N$ static cameras. Camera $n$ is parameterized by intrinsics $\bm{K}^n\in\mathbb{R}^{3\times3}$, camera-to-world rotation $\bm{R}^n\in SO(3)$, and translation $\bm{t}^n\in\mathbb{R}^3$, with pixel domain $\Pi\subset\mathbb{Z}^2$.

At each $k\in\mathcal{K}$, the observations from camera $n$ are an RGB image $I_k^n:\Pi\to[0,1]^3$, a depth map $D_k^n:\Pi\to\mathbb{R}_{\geq0}$, and a moving-object mask $M_k^n:\Pi\to\{0,1\}$. We denote the multi-view observations at time $k$ by
\[
\mathcal{O}_k=\{(I_k^n,D_k^n,M_k^n)\}_{n=1}^N,
\qquad \mathcal{O}_k\in\mathcal{O},
\]
where $\mathcal{O}$ denotes the space of multi-view observations.

We consider two prediction settings. In \emph{in-between prediction}, observations are sampled every $s\geq 2$ steps, with $\mathcal{K}=\{0,s,2s,\ldots\}$, and the targets are the $s-1$ unobserved steps between consecutive observations. In \emph{future prediction}, observations are available up to the middle of the clip, $\mathcal{K}=\{0,1,\ldots,k\}$, and the targets are the next $J=10$ steps, $\{k+1,\ldots,k+J\}$. At each observed step, all $N$ cameras provide observations.

\textbf{Volumetric Fusion of Multi-view Observations.} We construct a voxel representation of the scene in world coordinates by fusing observations from all cameras, following the volumetric fusion paradigm~\cite{curless1996volumetric,newcombe2011kinectfusion}. The grid covers the region in which the objects of a clip move, at a voxel size shared by all scenes, and its size does not depend on the number of cameras. Each voxel is occupied, free, or unseen.

Let $\Delta>0$ denote the voxel size and $\bm{o}\in\mathbb{R}^3$ the minimum corner of the grid in world coordinates, with $G_i$ voxels along axis $i$. We map a world point $\bm{y}\in\mathbb{R}^3$ to continuous grid coordinates using
\[
\bm{g}(\bm{y}) = \frac{\bm{y}-\bm{o}}{\Delta}-\frac{1}{2}.
\]
The grid domain is the box $\Omega=\prod_{i=1}^{3}[-\tfrac{1}{2},\,G_i-\tfrac{1}{2}]\subset\mathbb{R}^3$ in grid coordinates, the image of the grid's world-space box under $\bm{g}$. Its voxel centers form the index set $\Omega_\Delta=\Omega\cap\mathbb{Z}^3$.

Each voxel stores a six-dimensional feature comprising RGB color, objectness, occupancy, and visibility. The resulting fused state belongs to
\[
\mathcal{Z}=\{\bm{z}\mid \bm{z}:\Omega_\Delta\rightarrow\mathbb{R}^6\}.
\]
We obtain this state through a fusion map $\mathcal{F}:\mathcal{O}\rightarrow\mathcal{Z}$, computed directly from the observations and known camera parameters. Specifically, under the pinhole camera model~\cite{hartley2004multiple}, each pixel with valid depth is back-projected to a 3D point (\cref{eq:unproject}). Points from all views are then binned into the voxel grid to determine occupancy and to assign voxel features (\cref{eq:occupancy,eq:splat}), and each voxel center is tested against the depth maps to determine visibility.

For each camera $n$, a pixel $\bm{u}\in\Pi$ with valid depth is unprojected into world coordinates as
\begin{equation}
  \bm{y} = \bm{R}^n \big( D_k^n(\bm{u})\, (\bm{K}^n)^{-1} \tilde{\bm{u}} \big) + \bm{t}^n,
  \label{eq:unproject}
\end{equation}
where $\tilde{\bm{u}}\in\mathbb{R}^3$ is the homogeneous coordinate of the pixel center. Let $\mathcal{P}_k\subset\mathbb{R}^3$ denote the resulting point set from all cameras at step $k$. Each point $\bm{y}\in\mathcal{P}_k$ retains the color $\bm{\rho}_{\bm{y}}\in[0,1]^3$ and moving-object mask label $m_{\bm{y}}\in\{0,1\}$ of its source pixel.

A voxel $\bm{x}\in\Omega_\Delta$ is marked occupied if at least three points fall within it:
\begin{equation}
  q_k(\bm{x}) = \mathbf{1}\Big[\, \big|\{\bm{y}\in\mathcal{P}_k : \operatorname{round}(\bm{g}(\bm{y})) = \bm{x}\}\big| \geq 3 \,\Big].
  \label{eq:occupancy}
\end{equation}
At occupied voxels, we aggregate color and objectness by trilinear splatting of the points from all cameras:
\begin{equation}
  [\bm{\rho}_k(\bm{x}), o_k(\bm{x})]
  = \sum_{\bm{y}\in\mathcal{P}_k}
  \bar{w}(\bm{y},\bm{x})[\bm{\rho}_{\bm{y}},m_{\bm{y}}],
  \label{eq:splat}
\end{equation}
where $\bar{w}(\bm{y},\bm{x})=w(\bm{y},\bm{x})/\sum_{\bm{y}'\in\mathcal{P}_k}w(\bm{y}',\bm{x})$ normalizes the trilinear weight $w(\bm{y},\bm{x})=\prod_{i=1}^{3}\max\big(0,1-|g_i(\bm{y})-x_i|\big)$, so that $\sum_{\bm{y}\in\mathcal{P}_k}\bar{w}(\bm{y},\bm{x})=1$ at every occupied voxel. Occupancy thus assigns each point to its nearest voxel, while its color and mask label are spread over the eight voxels around it. Both quantities are set to zero at unoccupied voxels. We additionally define an unseen flag $\nu_k(\bm{x})\in\{0,1\}$. It is set at an unoccupied voxel when, in every camera, the voxel center projects outside the image or lies more than $\Delta/2$ behind the observed depth at its pixel. An unoccupied voxel with $\nu_k(\bm{x})=0$ lies in front of an observed surface along some camera ray, so it is observed free space. The fused state is then $\mathcal{F}(\mathcal{O}_k)=\bm{z}_k$, where, 
\[
\bm{z}_k(\bm{x})=
[\bm{\rho}_k(\bm{x}),o_k(\bm{x}),q_k(\bm{x}),\nu_k(\bm{x})]
\in\mathbb{R}^6.
\]
We extract the moving-object points as the occupied voxels with objectness above $0.5$:
\[
\mathcal{X}_k =
\{\bm{x}\in\Omega_\Delta:q_k(\bm{x})=1,\ o_k(\bm{x})>0.5\}.
\]
We use the depth maps and object masks provided by the datasets.

\subsection{Learning a Continuous-time Dynamics Operator}\label{sec:operator}

For a prediction interval beginning at observation $k_0$, we construct the context $\ctx=(\z_{k_0},\z_{k_p},\z_{k_1},e,s)$, for, $k_p=\max(k_0-s,0),\text{ and } k_1=k_0+s,$
where $s$ is the observation spacing and $e\in\{0,1\}$ indicates whether the succeeding observation is available. Thus, $\ctx\in\mathcal{C}=\mathcal{Z}^3\times\{0,1\}\times\mathbb{N}$. In in-between prediction, both bounding observations are available and $e=1$. In future prediction, $k_0=k$ is the last observed step, $k_p=k-2$, and $s=2$. The succeeding observation is unavailable, so we set $\z_{k_1}=\bm{0}$ and $e=0$.

We learn an operator $\mathcal{G}_\theta$ that maps this context to a time-dependent 3D velocity field:
\[
\mathcal{G}_\theta:\mathcal{C}\rightarrow\mathcal{V},
\qquad
\ctx\mapsto\vel_\theta(\cdot,\cdot\mid\ctx),
\]
where $\mathcal{V}=\{\vel:\Omega\times[0,H]\rightarrow\mathbb{R}^3\}$ with velocities in voxels per time step, and $H$ is the longest horizon the operator predicts. We use $H=12\geq\max(s,J)$: in-between prediction integrates over $t\in[0,s]$ and future prediction over $t\in[0,J]$, with $t$ measured from $k_0$.

We implement $\mathcal{G}_\theta$ as a 3D U-Net~\cite{ronneberger2015unet,cicek20163d} trained across scenes. In a single forward pass, the network outputs the velocity fields at all $H{+}1$ integer times $t=0,\ldots,H$, the node fields $\V_0,\ldots,\V_H$ of \cref{fig:pipeline}. Velocities at intermediate times are linearly interpolated between consecutive node fields (\appref{sec:x_method}). The inputs are the three fused states, $e$, and $s$, so time enters only as the offset $t$ from $k_0$, never as an absolute timestamp. The output layer is zero-initialized, so the initial velocity field is zero.

\subsection{Evolving the Dynamics through Advection}\label{sec:advect}

The predicted velocity field defines an ODE for the motion of each object point:
\begin{equation}
  \frac{d\bm{x}(t)}{dt} = \vel_\theta(\bm{x}(t),t\mid\ctx),
  \qquad \bm{x}(t_a)=\bm{x}_a.
  \label{eq:ode}
\end{equation}
We denote its solution at time $t_b$ by the flow map
$\Phi_{t_a\to t_b}^{\ctx}:\Omega\to\mathbb{R}^3$.
For an observed object point set $\X_{t_a}$, the predicted point set is
\[
  \hat{\X}_{t_b}
  = \{\Phi_{t_a\to t_b}^{\ctx}(\bm{x}):\bm{x}\in\X_{t_a}\},
\]
with points outside $\Omega$ discarded. Velocities are sampled trilinearly from the node fields, and intermediate integration stages that leave $\Omega$ use the velocity at the nearest grid boundary.

We solve \cref{eq:ode} with a second-order midpoint scheme that evaluates both stages at the substep's mid-time. Given $L$ substeps, let $\delta=(t_b-t_a)/L$, $t_l=t_a+l\delta$, and $\bm{x}^{(0)}=\bm{x}_a$. The update is
\begin{equation}
  \begin{aligned}
    \tilde{\bm{x}}^{(l)} &= \bm{x}^{(l)} + \tfrac{\delta}{2}\,\vel_\theta\big(\bm{x}^{(l)},\,t_l+\tfrac{\delta}{2}\mid\ctx\big),\\
    \bm{x}^{(l+1)} &= \bm{x}^{(l)} + \delta\,\vel_\theta\big(\tilde{\bm{x}}^{(l)},\,t_l+\tfrac{\delta}{2}\mid\ctx\big),
  \end{aligned}
  \label{eq:midpoint}
\end{equation}
for $l=0,\ldots,L-1$. The final position $\bm{x}^{(L)}$ approximates
$\Phi_{t_a\to t_b}^{\ctx}(\bm{x}_a)$. We increase $L$ with the integration interval $|t_b-t_a|$ (\cref{tab:appendix_hparams}). Since $\delta$ can be positive or negative, the same velocity field supports forward and backward integration.

For in-between prediction, we evaluate the operator once for each observation gap $(k_0,k_1)$, with time measured from $k_0$. Each target step is predicted by advecting points from the nearer bounding observation. For future prediction, we advect the points at the last observed step $k$ forward using $\ctx=(\z_k,\z_{k-2},\bm{0},0,2)$. Both cases are summarized by
\begin{equation}
  \begin{gathered}
  \hat{\X}_{k_0+j} =
  \begin{cases}
    \Phi_{0\to j}^{\ctx}(\X_{k_0}), & j < s/2,\\
    \Phi_{s\to j}^{\ctx}(\X_{k_1}), & j \geq s/2,
  \end{cases}\\
  \hat{\X}_{k+j} = \Phi_{0\to j}^{\ctx}(\X_k),
  \end{gathered}
  \label{eq:predict}
\end{equation}
where the first expression applies to in-between prediction for $j=1,\ldots,s-1$, with the middle step $j=s/2$ of an even gap advected from the succeeding observation, and the second to future prediction for $j=1,\ldots,J$.

Only moving-object points are advected. Static occupied voxels,
\[
  \mathcal{S} =
  \bigcup_{k\in\mathcal{K}}
  \{\bm{x}\in\Omega_\Delta:q_k(\bm{x})=1,\ o_k(\bm{x})\leq0.5\},
\]
remain fixed, and each takes its color from the latest observation in which it is occupied. To construct the predicted state $\hat{\z}\in\mathcal{Z}$, we round the advected points to voxel locations and retain their source colors and objectness values. We render the resulting points from arbitrary viewpoints over a background constructed from the observed images (\appref{sec:appendix_impl}, \cref{fig:pipeline}). The complete inference procedure is given in \appref{alg:x_inference}.

\begin{table*}[tb]
  \caption{\textbf{Comparison scenes.} F1 (\%, \up) of the predicted 3D object points. \best{Bold}: best, \second{underline}: second best.}
  \label{tab:perscene_3d}
  \small%
  \centering%
  \setlength{\tabcolsep}{1pt}%
\begin{tabular*}{\textwidth}{@{\extracolsep{\fill}}ll*{12}{c}@{}}
    \toprule
    & & \multicolumn{7}{c}{In-between prediction} & \multicolumn{5}{c}{Future prediction} \\
    \cmidrule(lr){3-9}\cmidrule(l){10-14}
    Method & Input
      & \shortstack{Elastic\\Fall} & \shortstack{DropIn\\Liquid} & \shortstack{Crank\\Push} & \shortstack{Panel\\Spin} & \shortstack{Double\\Pendulum} & \shortstack{Balloon\\Lift} & Mean
      & \shortstack{Elastic\\Fall} & \shortstack{Panel\\Spin} & \shortstack{Double\\Pendulum} & \shortstack{Balloon\\Lift} & Mean \\
    \midrule
    FreeGave & RGB, D\textsubscript{0}, M
      & 77.1 & 21.7 & \second{85.2} & 56.3 & 91.5 & 82.8 & 69.1 & 0.7 & 18.6 & 33.6 & 73.3 & 31.5 \\
    TRACE & RGB, D\textsubscript{0}, M
      & 0.0 & 15.1 & 70.2 & 32.5 & 15.7 & 45.8 & 29.9 & 2.5 & 32.8 & 34.4 & 44.0 & 28.4 \\
    SoM & RGB-D, M, 1 view
      & 77.5 & 49.1 & 73.5 & 72.2 & 82.2 & 49.3 & 67.3 & 13.5 & 23.8 & 44.0 & 43.5 & 31.2 \\
    MoSca & RGB-D, M, 1 view
      & 73.5 & 19.5 & 40.6 & 33.6 & 52.2 & 40.3 & 43.3 & 5.9 & 31.0 & 27.1 & 50.6 & 28.6 \\
    STv2 & RGB-D, M
      & \second{96.4} & \second{65.1} & 84.7 & \best{89.2} & \best{96.1} & \second{95.5} & \second{87.8} & 4.2 & \best{57.4} & 50.4 & \best{90.8} & \second{50.7} \\
    OpenD4RT & RGB-D, M
      & 93.2 & 61.6 & 82.2 & 77.3 & 94.5 & 95.3 & 84.0 & \second{16.5} & 30.7 & \second{63.4} & 73.1 & 45.9 \\
    \oursrow \ours{} & RGB-D, M
      & \best{97.3} & \best{77.9} & \best{87.2} & \second{86.3} & \second{95.4} & \best{95.8} & \best{90.0} & \best{22.0} & \second{47.6} & \best{80.7} & \second{83.2} & \best{58.4} \\
    \bottomrule
  \end{tabular*}
\end{table*}
\subsection{Learning Objective}\label{sec:training}

We train $\mathcal{G}_\theta$ by comparing advected object points with the object points reconstructed from the target observations. Since training clips provide observations at every time step, we obtain target point sets by fusing the depth maps and object masks from all cameras. No ground-truth velocities or ground-truth point correspondences are required. Instead, we compare the advected and target points as unordered sets, which avoids assigning persistent identities to surface points on deformable objects.

Each training sample comprises a context $\ctx$, a source point set $\mathcal{Q}=\X_{t_a}\subset\Omega$, and one or more target steps $j$. For each target, we advect the source points to obtain $\hat{\X}_j=\Phi_{t_a\to j}^{\ctx}(\mathcal{Q})$ and compare them with the corresponding target point set $\mathcal{Y}_j\subset\Omega$. We use the symmetric Chamfer distance in voxel coordinates:
\[
d(\mathcal{A},\mathcal{B}) =
\tfrac{1}{2}\Big(
\operatorname*{mean}_{\bm{a}\in\mathcal{A}}
\min_{\bm{b}\in\mathcal{B}}\|\bm{a}-\bm{b}\|
+
\operatorname*{mean}_{\bm{b}\in\mathcal{B}}
\min_{\bm{a}\in\mathcal{A}}\|\bm{b}-\bm{a}\|
\Big).
\]
To account for differences in motion magnitude across samples, we normalize the distance by the error of a static prediction, in which the source points remain at their initial positions:
\begin{equation}
  \ell_{\mathrm{point}} =
  \operatorname*{mean}_{(\mathcal{Q},j)}
  \frac{
    d(\hat{\X}_j,\mathcal{Y}_j)
  }{
    \max\big(d(\mathcal{Q},\mathcal{Y}_j),\tfrac{1}{2}\big)
  }.
  \label{eq:pointloss}
\end{equation}
The mean is taken over source sets and target steps in a batch. The denominator provides a sample-dependent scale, with a lower bound of $1/2$ to avoid instability when the target displacement is small. With the zero-initialized velocity field, the points stay in place and each term equals $\min\big(1,\,2\,d(\mathcal{Q},\mathcal{Y}_j)\big)$, so $\ell_{\mathrm{point}}\leq1$, with equality when every static error is at least half a voxel.

The Chamfer loss does not constrain the local geometry of the advected point set. We therefore add a local distance-preservation regularizer that penalizes changes in the distances between neighboring source points. For $\mathcal{Q}=\{\bm{x}_i\}_{i=1}^{|\mathcal{Q}|}$, let $\hat{\bm{x}}_i=\Phi_{t_a\to j}^{\ctx}(\bm{x}_i)$ and let $\mathcal{N}(i)$ contain the six nearest neighbors of $\bm{x}_i$ in the source set. We define
\begin{equation}
  \ell_{\mathrm{rigid}} =
  \operatorname*{mean}_{(\mathcal{Q},j)}
  \operatorname*{mean}_{i,\ n \in \mathcal{N}(i)}
  \left|
    \|\hat{\bm{x}}_i-\hat{\bm{x}}_n\|
    -
    \|\bm{x}_i-\bm{x}_n\|
  \right|.
  \label{eq:rigid}
\end{equation}
The training objective is
\[
\mathcal{L}(\theta)
=
\ell_{\mathrm{point}}+\lambda\,\ell_{\mathrm{rigid}},
\]
where $\lambda>0$ controls the regularization strength. We introduce the rigidity term only in the later stages of training.

We sample contexts and source-target intervals using both the spacings and horizons considered at test time and randomly sampled intervals. Training also uses a horizon curriculum. To avoid batches dominated by static examples, we resample a training window until it contains a sufficient number of moving object voxels. Further implementation details are provided in \appref{sec:appendix_impl}.

\section{Experiments}\label{sec:experiments}
\begin{figure*}[!tb]
  \centering
  \includegraphics[width=\linewidth]{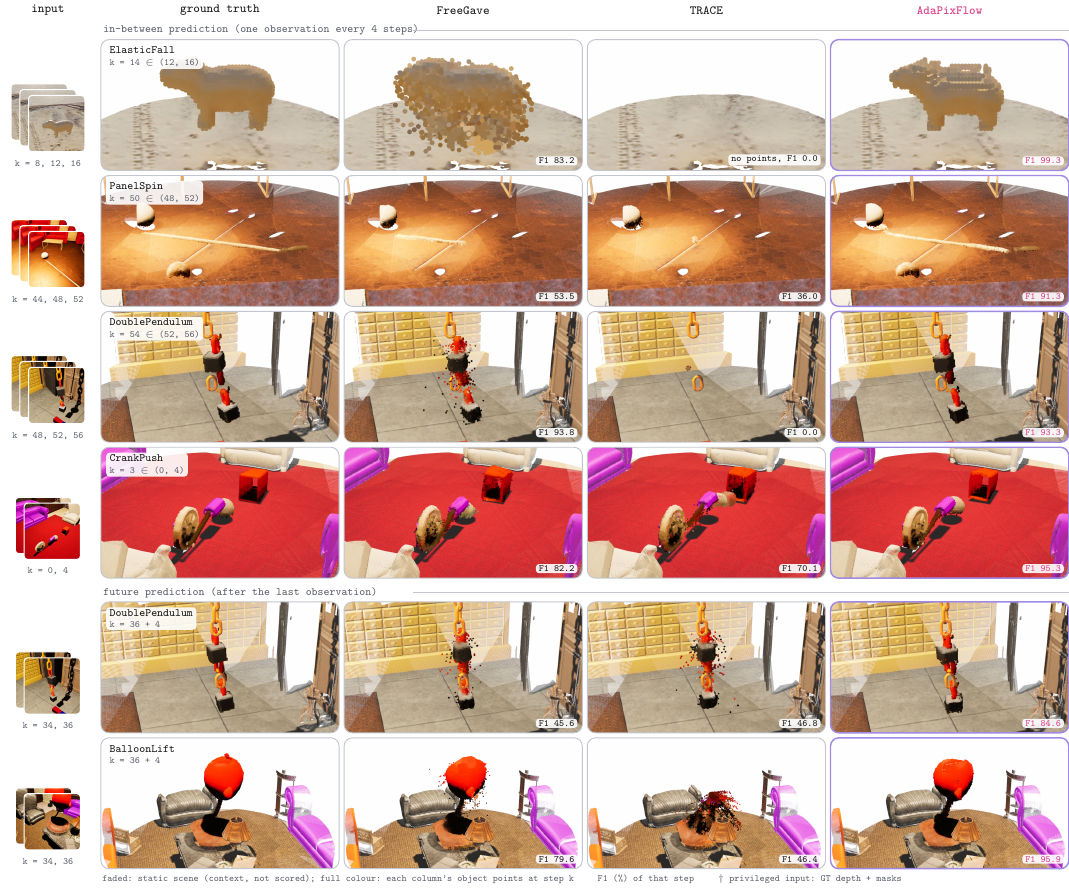}
  \caption{Predicted 3D object points for in-between (top) and future prediction (bottom) on the comparison scenes: input, ground truth, FreeGave, TRACE, and \ours{}, with each cell's F1 (\%).}
  \label{fig:qualitative}
\end{figure*}
\begin{table*}[tb]
  \caption{\textbf{Physics benchmarks.} F1 (\%, \up) on three benchmarks not used for training, Spring-Gaus~\cite{zhong2024reconstruction}, PhysGaia~\cite{kim2026physgaia}, and Kubric-4D~\cite{vanhoorick2024gcd,greff2022kubric}. Targets are pooled into linear and non-linear in-between motion and future prediction 4--10 steps ahead. \textsuperscript{\ddag}: excluded, as its F1 on the observed steps is below 90.}
  \label{tab:physbench}
  \small%
  \centering%
  \setlength{\tabcolsep}{1.5pt}%
\begin{tabular*}{\textwidth}{@{\extracolsep{\fill}}llrcccccc>{\columncolor{oursbg}}cc@{}}
    \toprule
    Dataset & Subset & $n$ & FreeGave & TRACE & SoM & MoSca & STv2 & OpenD4RT & \ours{} & Oracle \\
    & & & \multicolumn{4}{c}{\textit{per-scene fit}} & \multicolumn{2}{c}{\textit{Kubric-trained}} & \textit{zero-shot} & \textit{GT states} \\
    \cmidrule(lr){4-7}\cmidrule(lr){8-9}\cmidrule(lr){10-10}\cmidrule(l){11-11}
    \midrule
    Spring-Gaus & linear & 28 & 64.1 & 84.3 & 73.3 & 59.2 & \best{96.0} & 94.1 & \second{95.7} & 99.3 \\
                            & non-linear &  8 & 66.7 & 70.5 & 59.8 & 48.5 & 69.7 & \second{71.6} & \best{72.3} & 75.1 \\
                            & future, $j{\geq}4$ & 21 & 35.1 & 24.3 & \second{47.3} & 27.1 & 43.7 & 39.8 & \best{52.7} & 39.0 \\
    \midrule
    PhysGaia & linear & 12 & 55.3 & 54.4 & 62.1 & 35.6 & --\textsuperscript{\ddag} & \second{91.5} & \best{98.3} & 98.2 \\
                            & non-linear & 14 & 41.6 & 42.7 & 40.0 & 20.6 & --\textsuperscript{\ddag} & \second{60.1} & \best{61.4} & 64.6 \\
                            & future, $j{\geq}4$ &  7 & 56.1 & \best{67.8} & 48.0 & 36.4 & --\textsuperscript{\ddag} & \second{65.8} & 62.2 & 54.6 \\
    \midrule
    Kubric-4D & linear & 38 & 50.3 & 50.4 & 73.2 & 36.0 & \best{99.2} & 97.5 & \second{98.7} & 99.2 \\
                            & non-linear &  2 & 57.7 & 56.3 & 74.3 & 42.8 & \best{97.3} & \second{95.5} & 91.6 & 86.2 \\
                            & future, $j{\geq}4$ & 20 & 62.3 & 56.8 & 57.9 & 38.9 & \best{85.6} & 70.9 & \second{79.5} & 64.9 \\
    \bottomrule
  \end{tabular*}
\end{table*}

Our experiments address three questions. (Q1) How does \ours{} compare with methods that optimize scene-specific representations or reconstruct observed frames when predicting object states at unobserved times (\cref{sec:exp_perscene})? (Q2) For which types of motion does learning a dynamics operator provide the greatest benefit, particularly when predicting between observations and beyond the observed sequence (\cref{sec:exp_motion})? (Q3) Can the learned operator generalize to unseen scenes and datasets without per-scene optimization (\cref{sec:exp_crossscene})?

\textbf{Setup.} We evaluate on PhysInOne~\cite{zhou2026physinone}, a multi-view dataset of dynamic scenes designed to study physical phenomena. It covers 71 phenomena and provides RGB images, depth, and segmentation from 12--15 static calibrated cameras per scene. We exclude granular and light phenomena, hold out 10\% of the scenes in each of the five tiers of the Train split (continuum, rigid mechanics with one, two, or three phenomena, and invisible forces), and train on 152 of the remaining scenes. We evaluate on 15 held-out Test scenes, 8 Val scenes, and three physics benchmarks not used during training. All baselines are run on six comparison scenes (ElasticFall, DropInLiquid, CrankPush, PanelSpin, DoublePendulum, and BalloonLift), five of which are Test scenes and one, ElasticFall, a Val scene.

Each time step corresponds to two video frames (1/15\,s). Unless stated otherwise, in-between prediction uses an observation spacing of $s=4$. Views 0 and 1 are reserved for novel-view evaluation and are not provided to any method as input. All methods receive the calibrated camera parameters, and each receives the subset of the observed RGB images (RGB), depth maps (D), and moving-object masks (M) listed in \cref{tab:perscene_3d}. Paired RGB-D observations are standard in multi-view datasets and capture systems.

\textbf{Baselines.} We evaluate against two groups of methods. FreeGave~\cite{li2025freegave} and TRACE~\cite{li2025trace} are per-scene dynamics models, using their released settings for PhysInOne. For each scene and task, we fit them to the observed RGB views, using depth from the first frame (D\textsubscript{0}) and masks to select object Gaussians. Training is performed at the 280\,px scoring resolution, which may disadvantage both methods (\appref{sec:appendix_baselines}).

Shape of Motion (SoM)~\cite{wang2025shape}, MoSca~\cite{lei2025mosca}, SpatialTrackerV2 (STv2)~\cite{xiao2025spatialtrackerv2}, and OpenD4RT (an unofficial reproduction of D4RT~\cite{zhang2026efficiently}) receive the same depth and masks as \ours{}. Since these methods estimate states at observed frames, we interpolate or extrapolate their tracks to obtain predictions at the target times (\appref{sec:appendix_juryrig}). Uni4R~\cite{yang2026marrying} has no released implementation, so we study its formulation in \appref{sec:x_uni4r}.

\textbf{Metrics.} We compare predicted 3D object points against points reconstructed by fusing all views at each target time, including the held-out views. We report F1 at a 2.5\,cm threshold (two voxels), evaluated within a spatial window defined by the ground-truth motion over the clip. We score a target time only if the object has moved by more than one voxel since the most recent observation. Results in \cref{tab:perscene_3d} are averaged with equal weight per scene.

\begin{table}[tb]
  \caption{\textbf{Held-out scenes.} F1 (\%, \up) and Chamfer distance (CD, mm, \down) of \ours{} on the test and Val scenes.}
  \label{tab:crossscene}
  \small%
  \centering%
  \setlength{\tabcolsep}{2.6pt}%
\begin{tabular}{l cccccc cc}
    \toprule
     & \multicolumn{6}{c}{In-between} & \multicolumn{2}{c}{Future} \\
    \cmidrule(lr){2-7}\cmidrule(lr){8-9}
     & \multicolumn{2}{c}{$s{=}2$} & \multicolumn{2}{c}{$s{=}4$} & \multicolumn{2}{c}{$s{=}8$} & \multicolumn{2}{c}{10 steps} \\
    \cmidrule(lr){2-3}\cmidrule(lr){4-5}\cmidrule(lr){6-7}\cmidrule(lr){8-9}
    Scenes & F1 & CD & F1 & CD & F1 & CD & F1 & CD \\
    \midrule
    Test (15) & 90.9 & 10.4 & 84.2 & 15.8 & 78.7 & 24.0 & 68.3 & 41.5 \\
    Val (8)   & 99.9 & 4.3  & 97.6 & 5.2  & 90.8 & 9.2  & 55.8 & 42.3 \\
    \bottomrule
  \end{tabular}
\end{table}

\begin{table}[t]
  \caption{\textbf{Rendered predictions.} Chamfer distance (mm) and image metrics on the held-out views. Marks as in \cref{tab:perscene_3d}.}
  \label{tab:perscene_img}
  \small
  \centering
  \setlength{\tabcolsep}{3.2pt}
  \begin{tabular}{@{}lccccc@{}}
    \toprule
    Method & CD \down & IoU \up & PSNR \up & SSIM \up & LPIPS \down \\
    \midrule
    \multicolumn{6}{@{}l}{\textit{In-between prediction}} \\
    FreeGave & \second{28.7} & \second{0.815} & \best{21.45} & \second{0.786} & \second{0.249} \\
    TRACE & 96.9 & 0.489 & 19.76 & 0.716 & 0.340 \\
    \oursrow \ours{} & \best{6.0} & \best{0.828} & \second{21.22} & \best{0.801} & \best{0.225} \\
    \midrule
    \multicolumn{6}{@{}l}{\textit{Future prediction}} \\
    FreeGave & \second{65.7} & \second{0.388} & \best{20.37} & \second{0.790} & \second{0.250} \\
    TRACE & 131.7 & 0.286 & 19.74 & 0.730 & 0.333 \\
    \oursrow \ours{} & \best{36.9} & \best{0.526} & \second{19.96} & \best{0.807} & \best{0.226} \\
    \bottomrule
  \end{tabular}
\end{table}

\begin{figure}[t]
  \centering
  \includegraphics[width=\linewidth]{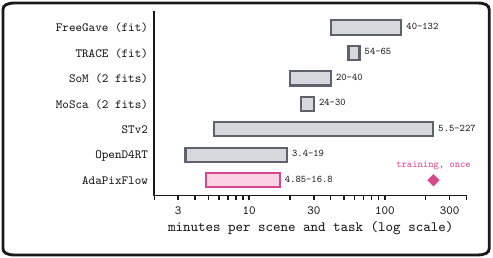}
  \caption{\textbf{Compute.} Minutes per scene and task on an RTX 5080 (log scale). SoM and MoSca include two candidate fits. \ours{} covers one scene and both tasks, end to end, and $\Diamond$ marks its one-off training.}
  \label{fig:x_compute}
\end{figure}

\subsection{Predicting States at Unobserved Times}\label{sec:exp_perscene}
\ours{} predicts object states more accurately between observations and beyond the observed sequence than the reconstruction and tracking baselines (\cref{tab:perscene_3d}, \cref{fig:qualitative}). The improvement is particularly pronounced in future prediction, where targets extend up to ten steps beyond the last observation. Quantitatively, \ours{} achieves mean F1 scores of 90.0 in-between and 58.4 in future prediction, compared with 87.8 and 50.7 for STv2, the next-best method.

STv2 and OpenD4RT are the closest baselines for in-between prediction. Both track points across observed frames without per-scene optimization, and STv2 outperforms \ours{} on PanelSpin and DoublePendulum. In future prediction, STv2 is slightly more accurate over the first three steps, but \ours{} leads from step four onward (\appref{tab:juryrig_split}). Extrapolating tracks at constant velocity drifts away from curved and accelerating motion. \ours{} predicts all ten future steps from one forward pass. We examine chained prediction passes in \appref{sec:x_exp_horizon}.

The per-scene dynamics models perform substantially worse despite 0.7--2.2\,h of optimization per scene (\cref{fig:x_compute}). Their RGB-based reconstructions miss substantial portions of the object even at observed frames, whereas our depth-based fusion provides more complete geometry (\appref{sec:appendix_baselines}).

\subsection{Motion Underdetermined by the Observations}\label{sec:exp_motion}
We examine whether the advantage of \ours{} depends on motion complexity. Using ground-truth states, we divide target times into linear and non-linear subsets according to whether an oracle linear interpolation achieves an F1 of at least 90. For each in-between target, the oracle has access to the exact object states at both ends of the observation gap and selects the best of three interpolation rules: centroid translation, rigid motion with interpolated rotation, or straight-line point trajectories. For future prediction, we use constant-velocity extrapolation from the last two observed states and evaluate targets 4--10 steps ahead. F1 is pooled over all targets in each subset.

On linear targets, \ours{} matches or outperforms the strongest trackers on both PhysInOne (\appref{tab:juryrig_split}) and the physics benchmarks (\cref{tab:physbench}). Its advantage is clearer on non-linear motion: it achieves the highest F1 on non-linear in-between targets across PhysInOne, Spring-Gaus, and PhysGaia, and on future targets 4--10 steps ahead in PhysInOne and Spring-Gaus. This is consistent with how the trackers are applied to unobserved times: we interpolate and extrapolate their tracks linearly in time (\appref{sec:appendix_juryrig}).

\subsection{Generalization to Unseen Scenes and Datasets}\label{sec:exp_crossscene}\label{sec:exp_physbench}
On held-out PhysInOne scenes (\cref{tab:crossscene}), prediction accuracy decreases as the observation spacing increases, while remaining high on the Val scenes despite their differences from the training distribution. On the 15 Test scenes, in-between F1 decreases from 90.9 to 78.7 as the spacing increases from 2 to 8 steps. The 8 Val scenes achieve comparable or higher scores (90.8--99.9), despite having backgrounds and a phenomenon (PlasticineFall) absent from training. Future prediction is less accurate on Val than on Test (55.8 vs. 68.3), as the Val forecast windows include ground contacts.

We further evaluate \ours{} without retraining on Spring-Gaus~\cite{zhong2024reconstruction}, PhysGaia~\cite{kim2026physgaia}, and Kubric-4D~\cite{vanhoorick2024gcd,greff2022kubric}. These datasets differ from PhysInOne in object materials, camera configurations, and scene scale. Kubric-4D provides depth and masks. For Spring-Gaus and PhysGaia, we render both from the provided particle data. Across scenes evaluated by all methods (\appref{tab:physbench_base}), \ours{} achieves the highest accuracy on Spring-Gaus in both tasks and on PhysGaia in-between. On Kubric-4D, it ranks second to STv2, which was trained on Kubric data.

Finally, \ours{} retains much of its performance with estimated rather than dataset-provided depth and masks. Using depth from Depth Anything~3~\cite{lin2026depth} and masks from SAM~2.1~\cite{ravi2025sam}, it achieves F1 scores of 78.5 in-between and 53.3 in future prediction on the comparison scenes. FreeGave, given the same masks, scores 66.5 and 30.0 (\appref{sec:x_exp_input}).

\section{Conclusion}\label{sec:conclusion}
\ours{} demonstrates that a shared continuous-time dynamics operator can predict 3D object motion across scenes from sparse multi-view observations. By combining calibrated geometric reconstruction with a learned velocity field, it predicts states at unobserved times without fitting a separate model to each scene. Experiments on held-out PhysInOne scenes and external physics benchmarks show competitive or improved prediction accuracy over reconstruction, tracking, and per-scene dynamics baselines, with the largest gains for non-linear motion and longer forecasts. These results suggest that learning motion in a shared 3D space provides a useful alternative to interpolating reconstructed trajectories.

\textbf{Future work.} The current operator models kinematics. Conditioning it on contact and force models would extend it to interactions such as collisions and bounces. Extending the forecast horizon through training on chained predictions is another direction. Finally, applying \ours{} to real multi-view captures will require handling errors in estimated depth and segmentation.

\FloatBarrier
{
    \small
    \bibliographystyle{ieeenat_fullname}
    \bibliography{references}
}
\section*{LLM Usage Statement}
We used a large language model as a coding and writing assistant: it wrote and ran code for data processing, baseline adaptation, evaluation, and figures under our direction, and edited the text. The authors designed the method and experiments, checked the code and every reported result, and take full responsibility for the paper.

\clearpage
\setcounter{page}{1}
\maketitlesupplementary
\renewcommand{\topfraction}{0.9}\renewcommand{\bottomfraction}{0.8}\renewcommand{\textfraction}{0.05}\renewcommand{\floatpagefraction}{0.6}
\setcounter{topnumber}{4}\setcounter{bottomnumber}{4}\setcounter{totalnumber}{8}\setcounter{dbltopnumber}{3}\renewcommand{\dbltopfraction}{0.9}\renewcommand{\dblfloatpagefraction}{0.6}
\appendix
\crefalias{section}{appendix}

\section{Additional Results}\label{sec:x_results}\label{sec:x_experiments}
In this section, we report the additional results referenced in the main text. We describe the settings needed to reproduce them in \cref{sec:appendix_impl,sec:appendix_eval,sec:appendix_baselines,sec:appendix_physbench}. All results come from the single model trained in the main text.

\subsection{Comparison Scenes}\label{sec:x_exp_perscene}
\begin{table}[!htbp]
  \caption{\textbf{Adapted 4D methods, Chamfer distance.} Median Chamfer distance (mm, \down) of every method on the comparison scenes, averaged per task. TRACE in-between covers five scenes.}
  \label{tab:juryrig_f1ch}
  \centering
  \small
  \setlength{\tabcolsep}{3pt}%
\begin{tabular}{@{}lcc@{}}
    \toprule
    Method & In-between & Future \\
    \midrule
    FreeGave~\cite{li2025freegave} & 28.7 & 65.7 \\
    TRACE~\cite{li2025trace} & 96.9 & 131.7 \\
    SoM~\cite{wang2025shape} & 16.9 & 56.4 \\
    MoSca~\cite{lei2025mosca} & 38.5 & 58.2 \\
    STv2~\cite{xiao2025spatialtrackerv2} & 6.4 & 48.9 \\
    OpenD4RT & 8.8 & 50.0 \\
    \oursrow \ours{} & 6.0 & 36.9 \\
    \bottomrule
  \end{tabular}%
\end{table}
\begin{figure*}[!t]
  \centering
  \includegraphics[width=\linewidth]{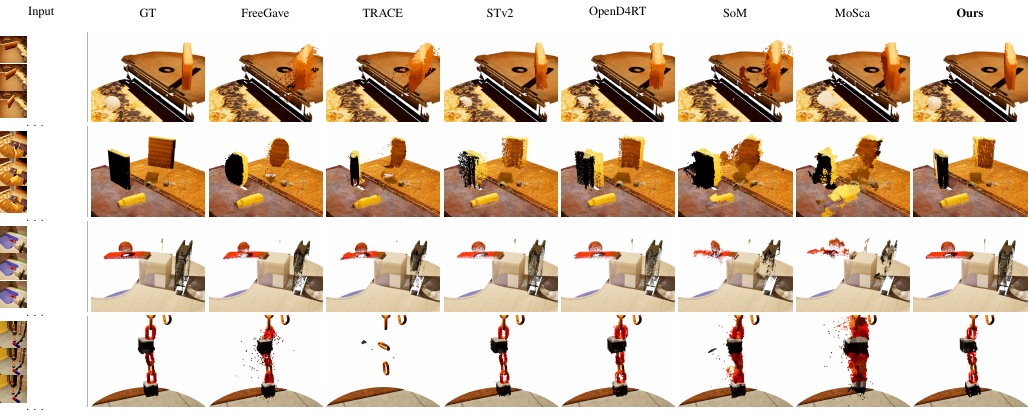}
  \caption{Predicted 3D object points of every method at one unobserved step of four held-out test scenes (in-between, $s = 4$), over the static scene.}
  \label{fig:qual_compare}
\end{figure*}

We report all candidates of the adapted 4D methods in \cref{tab:juryrig_main,tab:juryrig_f1ch}, and the subsets of the motion split in \cref{tab:juryrig_split}. \Cref{fig:qual_compare} shows every method on four held-out test scenes. To avoid selecting steps by their results, we always show the middle scored step, index $\lfloor (n-1)/2 \rfloor$ of the $n$ scored steps.

We find that the renderer does not affect our image metrics. When we render the fused state of the target step itself, as shown in the Voxel ceiling in \cref{tab:x_img_full}, we reach 33.91\,dB on training views in-between, so the remaining error comes from the predicted motion.

The oracle linear interpolation defines the motion split. However, we do not consider it an upper bound. STv2 exceeds it on non-linear in-between targets (78.3 vs.\ 69.9) and on targets 4--10 steps ahead (44.4 vs.\ 34.0). STv2 is strongest on PanelSpin. Without that scene, \ours{} leads STv2 on non-linear targets by 11.0 points (75.7 vs.\ 64.7), compared with 2.3 points when it is included (80.6 vs.\ 78.3).

The qualitative results show where the baselines fail. The tracks of STv2 lose correspondence on the liquid surface of DropInLiquid, and constant-velocity extrapolation carries objects through the floor in ElasticFall and along the tangent of the swing in DoublePendulum. OpenD4RT drops to 62.1 in-between when it uses its own geometry instead of our fusion. SoM and MoSca are fit to a single view, so they miss the unseen side of the objects. Finally, TRACE does not fit ElasticFall in-between (F1 of 0.0) and fits DoublePendulum poorly, although it fits both scenes with the inputs of the future task. We cannot tell whether this comes from our time scaling for in-between prediction or from the sparser observations (\cref{sec:appendix_baselines}).

\subsection{Held-out Scenes}\label{sec:x_exp_crossscene}
\begin{figure}[!htbp]
  \centering
  \includegraphics[width=\linewidth]{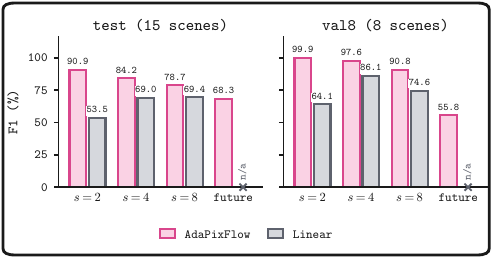}
  \caption{\textbf{Held-out scenes with references.} F1 (\%) of \ours{} and the Linear reference by spacing. Test means cover the scenes with scored steps.}
  \label{fig:x_crossscene}
\end{figure}
\begin{figure*}[!t]
  \centering
  \includegraphics[width=\linewidth]{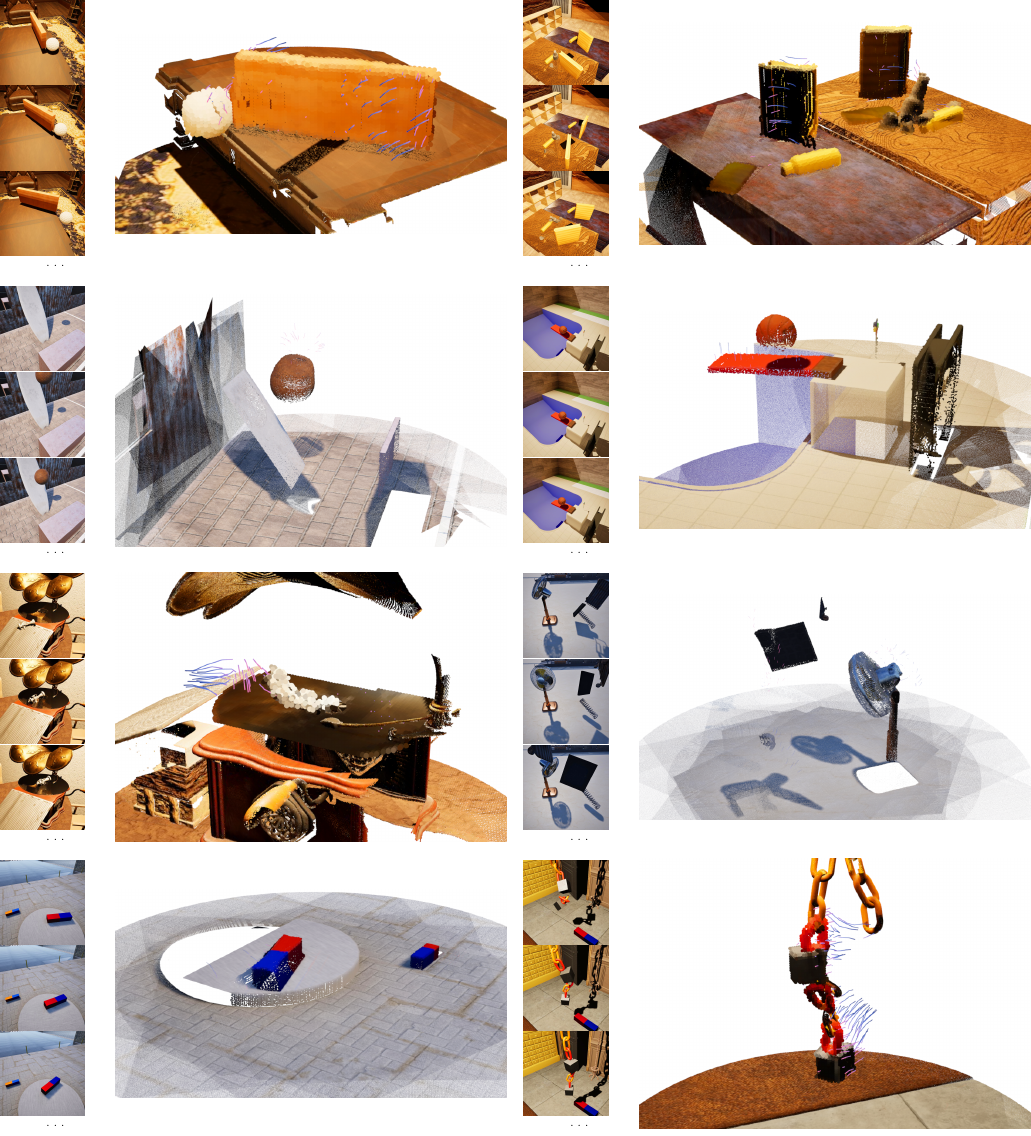}
  \caption{\ours{} on held-out PhysInOne test scenes: observed frames (left) and the predicted 3D object points at an unobserved step over the static scene, with pathlines of the six preceding steps (right).}
  \label{fig:qual_gallery}
\end{figure*}

\Cref{fig:x_crossscene} adds the Linear reference to \cref{tab:crossscene}, and \cref{fig:qual_gallery} shows eight held-out test scenes at the middle of the gap with the largest ground-truth change. We draw the pathlines only for display.

Linear blends the two observations per voxel, which fades objects instead of moving them. At the middle of a gap, a voxel occupied in only one observation keeps half its occupancy and falls below the threshold, so Linear recovers only 29\% of the ground-truth object voxels at $s{=}2$.

Accuracy varies across tiers. Tier B2 drops to 63.7 at $s{=}8$ and to 32.0 in future prediction, and two of its three scenes combine a spinning object with a collision.

The test scenes share their phenomena, and 14 of the 15 share a background, with training scenes. The Val scenes are therefore our strongest test of generalization: their backgrounds never appear in training, and neither does PlasticineFall, one of their two phenomena. Both phenomena are falling deformable objects, and we leave other unseen phenomena and real captures to future work. Finally, the window of this evaluation is capped at 128 voxels, which removes 22--56\% of the ground-truth object in DropInLiquid, CrankPush, and PanelSpin.

\subsection{Physics Benchmarks}\label{sec:x_exp_physbench}
\begin{figure*}[!t]
  \centering
  \includegraphics[width=\linewidth]{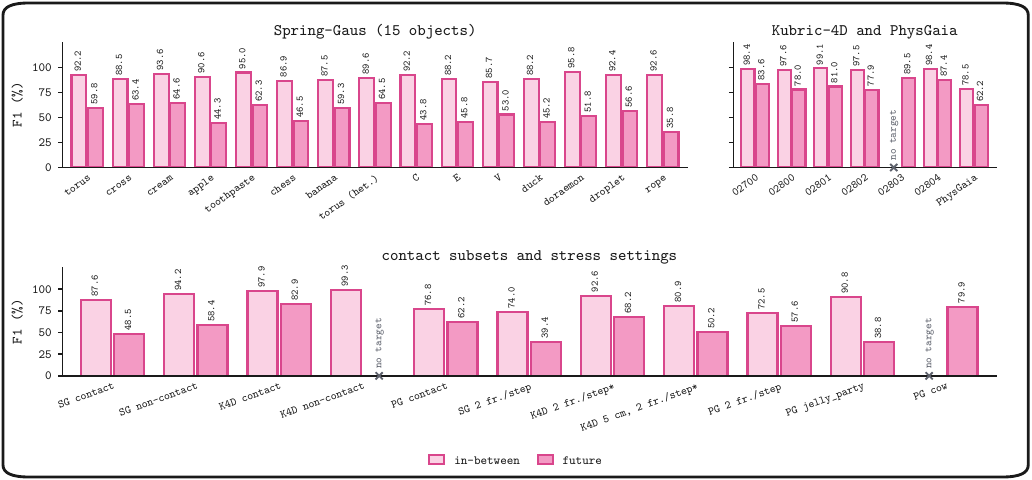}
  \caption{\textbf{Physics benchmarks, per scene and subset.} F1 (\%) of \ours{} (zero-shot), in-between (light) and future (dark), per scene, contact subset, and stress setting (two frames per step or 5\,cm voxels). $^\ast$: scn02700 and scn02800 only.}
  \label{fig:x_bench}
\end{figure*}

\Cref{tab:physbench_base} gives the scene means of every method, and \cref{fig:x_bench} breaks down the results of \ours{} by scene, contact subset, and stress setting. Averaged over the common subset, \ours{} is the most accurate method on Spring-Gaus in both tasks (90.5 and 55.8) and on PhysGaia in-between (78.5). On Kubric-4D, it ranks second (98.4 and 80.9) behind STv2 (99.2 and 86.3), which, like D4RT, was trained with Kubric data. We exclude STv2 on PhysGaia because its F1 on the observed steps in future prediction (75.4) is below the threshold of 90 that we fixed before scoring any target. These benchmarks also test extrapolation in the per-step dynamics, since gravity, measured in voxels per time step, is 1.1--9 times that of PhysInOne.

\subsection{Long-horizon Forecasting}\label{sec:x_exp_horizon}
\begin{figure}[!htbp]
  \centering
  \includegraphics[width=\linewidth]{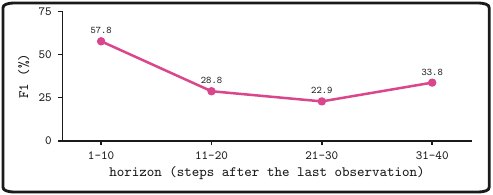}
  \caption{\textbf{Chained forecasts.} F1 (\%) by horizon on the five comparison scenes with scored future steps (all-view input, capped window). The last bin covers four scenes.}
  \label{fig:x_horizon}
\end{figure}
To forecast beyond 10 steps, we chain 10-step passes and condition each pass on the last two predicted states, two steps apart. Accuracy drops beyond 10 steps (\cref{fig:x_horizon}), whereas a pass restarted from an observed state stays accurate. This is expected, since the operator is trained on observed rather than predicted states, and it is why we evaluate forecasts up to 10 steps in the main text. We verified the chaining itself: its first pass equals the 10-step forecast, and a pass restarted from the ground-truth state stays within 0.4--4 voxels on ElasticFall in mid-flight. Two scenes account for the drop. In ElasticFall, the flow continues through the floor after the bounce (16 voxels of error at step 9 against 11 voxels of motion), and in the chaotic DoublePendulum, the centroid error reaches 34 voxels beyond 10 steps against 27 voxels of motion. Since the window of \cref{fig:x_horizon} differs from that of \cref{tab:perscene_3d}, its first bin is not comparable with the main results.

\subsection{Input Fusion}\label{sec:x_exp_input}
\begin{figure*}[!t]
  \centering
  \includegraphics[width=\linewidth]{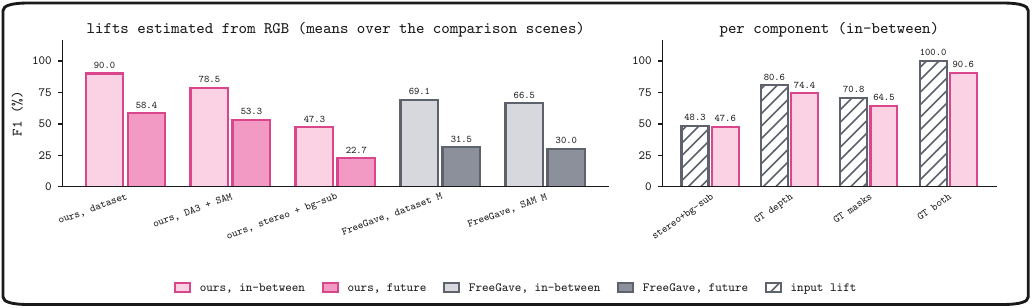}
  \caption{\textbf{Input fusion.} Left: means over the comparison scenes with fused inputs estimated from RGB, in-between (light) and future (dark). Right: the input fusion and \ours{} per component, in-between (GT: the dataset's).}
  \label{fig:x_lift}
\end{figure*}
\begin{figure}[!htbp]
  \centering
  \includegraphics[width=\linewidth]{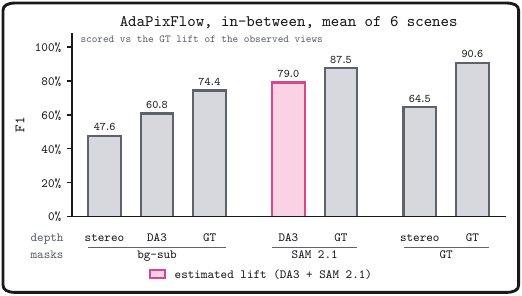}
  \caption{In-between F1 (\%, \up) of \ours{} with seven input fusions, grouped by masks and ordered by depth. Pink: estimated fusion.}
  \label{fig:inputs_chart}
\end{figure}

Here, we feed the same trained model fused inputs whose depth or masks are estimated from the RGB images of the observed views (settings in \cref{tab:x_lifts}). We keep the grid extent and the evaluation window of the main experiments. In \cref{fig:x_lift} (left), FreeGave reads the depth of the first frame, and the masks only select its object Gaussians. For the stereo fusion in future prediction, the labels also use the stereo depth of the first predicted time step, which gives them information from one unobserved step. We score the right panel of \cref{fig:x_lift} and \cref{fig:inputs_chart} against the fusion of the observed views, so the dataset fusion reaches 90.6 there instead of 90.0.

With the estimated fusion, using Depth Anything~3 depth and SAM~2.1 masks, \ours{} stays above FreeGave labeled with the same masks in both tasks. Depth is the most influential input. Stereo fails on the liquid of DropInLiquid and on the textureless balloon of BalloonLift (1.6k of 64.5k and 8.7k of 14.7k object voxels per frame), and replacing it raises the mean from 47.6 to 74.4. Replacing the labels has the next largest effect (47.6 to 64.5), since background subtraction misses the slow parts of CrankPush and over-labels the volume swept by the panel in PanelSpin. Finally, \ours{} stays below the accuracy of its own input fusion in every row, because the flow can only move the object points that the fusion contains.

\section{Method Details}\label{sec:x_method}
In this section, we describe the parts of the method that we omit from the main text.

\textbf{Velocity field.}
In one forward pass, $\mathcal{G}_\theta$ outputs $H{+}1 = 13$ node fields $\V_m : \Omega_\Delta \to \R^3$, where $\V_m$ is the velocity $m$ steps after the initial observation. We obtain the velocity at any position and time by interpolating between them,
\begin{equation}
  \vel_\theta(\bm{x}, t \mid \ctx) = (1 - a)\, \V_m(\bm{x}) + a\, \V_{m+1}(\bm{x}),
  \label{eq:x_velocity}
\end{equation}
where $m = \min(\lfloor t \rfloor, H - 1)$ and $a = t - m$. We sample each $\V_m$ trilinearly at off-grid positions, so the field is continuous in space and piecewise linear in time. In all our evaluations, query times are integer steps, and we integrate with 8 substeps per time step.

\begin{algorithm}[!htbp]
  \caption{Inference of \ours{} for in-between and future prediction.}
  \label{alg:x_inference}
  \begin{algorithmic}[1]
    \Require fused observations $\{\z_k\}_{k \in \mathcal{K}}$, trained network $\mathcal{G}_\theta$
    \Statex \textit{In-between prediction, observations every $s$ time steps:}
    \For{each gap $(k_0, k_1 = k_0 + s)$ with $k_0, k_1 \in \mathcal{K}$}
      \State $k_p \gets \max(k_0 - s, 0)$, \quad $\ctx \gets (\z_{k_0}, \z_{k_p}, \z_{k_1}, 1, s)$
      \State $\{\V_m\}_{m=0}^{H} \gets \mathcal{G}_\theta(\ctx)$ \Comment{one pass per gap}
      \For{$j = 1, \dots, s-1$}
        \If{$j < s/2$}
          \State $\hat{\X} \gets \Phi_{0 \to j}(\X_{k_0})$ \Comment{forward, \cref{eq:midpoint}}
        \Else
          \State $\hat{\X} \gets \Phi_{s \to j}(\X_{k_1})$ \Comment{backward}
        \EndIf
        \State $\hat{\z}_{k_0 + j} \gets$ \textsc{Voxelize}$(\hat{\X})$ over $\mathcal{S}$
      \EndFor
    \EndFor
    \Statex \textit{Future prediction, last observation $k$, $J$ query steps:}
    \State $\ctx \gets (\z_{k}, \z_{k-2}, \bm{0}, 0, 2)$, \quad $\{\V_m\}_{m=0}^{H} \gets \mathcal{G}_\theta(\ctx)$
    \For{$j = 1, \dots, J$}
      \State $\hat{\z}_{k + j} \gets$ \textsc{Voxelize}$(\Phi_{0 \to j}(\X_{k}))$ over $\mathcal{S}$
    \EndFor
    \State \textbf{return} predicted states $\hat{\z}$, rendered to any view (\appref{sec:appendix_impl})
  \end{algorithmic}
\end{algorithm}

\textbf{Inference.}
\Cref{alg:x_inference} summarizes inference. The first gap of in-between prediction has no preceding observation, so we set $k_p = k_0$ there. Within a gap, the forward and backward flows share one velocity field, which sees both neighboring observations, and we predict each step in a single direction.

\textbf{Predicted states and images.}
We round the advected points to the nearest voxel and drop the points that leave the evaluation window. Each predicted voxel carries the color of its source voxel, with occupancy and objectness set to one and the unseen flag to zero, and we place these voxels over the static voxels $\mathcal{S}$, where the latest observed color wins.
To render images, we work with pixels rather than voxels. We back-project the object pixels of all observed views from their depth at the observation, advect them along the same velocity field, and render them as points into the target camera at twice the image resolution. A z-buffer averages the points near the front surface, weighted by the angle between the source and target viewing directions, so that a training view mostly keeps its own pixels. The covered fraction of each pixel gives the object opacity $A$, and $A > 0.5$ defines the silhouette. We composite the object color $\hat{C}$ over a static image $B$ built only from observed data, $I = \hat{C} A + B (1 - A)$. For training views, $B$ takes at each pixel the observed image nearest in time in which the pixel is not part of an object. For novel views, we reproject the non-object depth points of the training views with a z-buffer and fill holes by normalized convolution. Points behind the static depth are hidden. \Cref{tab:x_impl} lists the settings.

\textbf{Training.}
We build the training targets by fusing all cameras of a training scene, 12, or 15 for special materials, and use them only as unordered object points per step. The network never sees a fusion from fewer views or from estimated depth during training, so such test inputs differ from the training distribution. We draw source points from the initial observation for forward targets and, when the next observation is given, from that observation for backward targets. Advected points outside the crop are left out of the prediction-to-target term of the Chamfer distance, and we do not supervise color. We found the redraw of training windows necessary: without it, a preliminary run on the same scenes collapsed to predicting no motion by step 3,000.

\textbf{Design choices.}
Our design follows from preliminary experiments with earlier models. Blending fused states per voxel, or interpolating values at fixed positions, cross-faded objects instead of moving them. Even a continuous-time operator trained on spline paths in a fused feature space reached only the IoU of per-voxel blending (0.79). Rolling out from the first observation compounded errors, reaching an IoU of 0.34 over 22 intervals, although each interval alone reached 1.00 from its true start. We therefore move points, start each prediction from an observation, and condition the field on both neighboring observations.


\section{Implementation and Reproducibility}\label{sec:appendix_impl}
We list the settings of \ours{} in \cref{tab:x_impl,tab:appendix_hparams}, and those of the fused inputs estimated from RGB in \cref{tab:x_lifts}.

\section{Evaluation Protocol}\label{sec:appendix_eval}
We list the evaluation protocol in \cref{tab:x_eval} and the scored steps of each comparison scene in \cref{tab:appendix_scenes}. We derive the grid extent of our fused states and the evaluation window from the ground truth of all views and time steps, but no method receives the images or depth of the novel views.

\section{Baselines}\label{sec:appendix_baselines}
\Cref{tab:x_baselines} lists the settings of the per-scene dynamics baselines, for which we use the released PhysInOne adaptations. We compare the compute of every method in \cref{fig:x_compute} of the main text.

\subsection{Adapting 4D Reconstruction and Tracking Methods}\label{sec:appendix_juryrig}
None of SoM~\cite{wang2025shape}, MoSca~\cite{lei2025mosca}, STv2~\cite{xiao2025spatialtrackerv2}, and D4RT~\cite{zhang2026efficiently} predicts unobserved time steps, so we adapt them as listed in \cref{tab:juryrig_took}. D4RT releases no weights, so every D4RT number comes from OpenD4RT, an unofficial reproduction.

For transparency, we note four departures from this protocol. First, we registered the stride-2 candidates of STv2 after its stride-4 target scores existed. With stride 4 everywhere, its means are 87.2 and 50.0 instead of 87.8 and 50.7. Second, the candidate set of OpenD4RT for its first scene, ElasticFall in-between, grew after we had seen target scores. Third, MoSca ran out of memory on three scene--task pairs and used the pre-registered fallback settings for short clips. Finally, selecting candidates on the observed steps does not always pick the better candidate on the targets. On ElasticFall future, two unselected candidates score 57.5 and 35.8, above \ours{}.

\subsection{Uni4R's Formulation under Sparse Observations}\label{sec:x_uni4r}
Uni4R~\cite{yang2026marrying} releases no code, so we test its formulation with a reduced reimplementation rather than the method itself.
Following its Sec.~3, we treat every pixel of an input frame as an anchor point. We interpolate features along a straight line between consecutive input frames, decode a per-pixel 3D velocity from them and the query time, and integrate it with Euler steps of 0.1 from the preceding frame. We train with its integral-consistency and flow-matching losses. Our model is smaller than Uni4R. It uses a DINOv2 ViT-S/14 encoder~\cite{oquab2024dinov2}, has 54\,M parameters in total, and is trained for 4.5\,h on one GPU on 118 Kubric-4D training scenes, whose rigid object poses give the ground-truth 3D position of every pixel at every frame. Like the other adapted methods, it reads the depth of the observed views, from which we obtain its anchor points.

We train it twice with the same budget, once on consecutive frames only, as Uni4R does, and once on inputs spaced 1 to 8 steps apart, as \ours{} does. We evaluate both on the Kubric-4D scenes of \cref{sec:appendix_physbench}. Each training view is a separate input, and we unite the points of all views and score them as in \cref{tab:x_eval}, averaging over the scenes with scored targets (five at $s{=}4$ and six otherwise).

When trained on consecutive frames, the model barely moves the points between sparse observations. It improves on holding the preceding observation, from which it integrates, by 6.4 F1 points at a spacing of 4 and by 2.8 at a spacing of 8, compared with 12.8 and 9.2 when trained on sparse spacings (\cref{tab:x_uni4r}). \ours{} integrates from the nearer observation and improves on holding it by 8.7 and 8.9 points. Beyond the last frame, where we extend Uni4R's feature interpolation past the last input, neither variant improves on holding the last observation by more than 0.4 points, whereas \ours{} improves on it by 12.4. Since our reimplementation is much smaller than Uni4R, these numbers describe how its formulation responds to sparse sampling, not Uni4R's own accuracy.

\begin{table}[!htbp]
  \caption{\textbf{Uni4R's formulation on Kubric-4D.} F1 (\%, \up). Top: our reduced reimplementation of Uni4R's formulation. Bottom: \ours{}. Each includes holding its starting observation.}
  \label{tab:x_uni4r}
  \small
  \centering
  \begin{tabular}{@{}lccc@{}}
    \toprule
    & \multicolumn{2}{c}{In-between} & Future \\
    \cmidrule(lr){2-3}\cmidrule(l){4-4}
    & $s{=}4$ & $s{=}8$ & 10 steps \\
    \midrule
    \multicolumn{4}{@{}l}{\emph{Uni4R formulation, reimplemented}} \\
    \quad trained on consecutive frames & 77.4 & 67.9 & 70.9 \\
    \quad trained on spacings 1--8 & 83.8 & 74.3 & 59.7 \\
    \quad holding its starting observation & 71.0 & 65.1 & 70.5 \\
    \midrule
    \multicolumn{4}{@{}l}{\emph{\ours{}}} \\
    \oursrow \quad \ours{} & 98.2 & 92.6 & 82.9 \\
    \quad holding its starting observation & 89.5 & 83.7 & 70.5 \\
    \bottomrule
  \end{tabular}
\end{table}

\section{Physics Benchmarks}\label{sec:appendix_physbench}
We list the setup of the physics benchmarks in \cref{tab:x_bench}. As on PhysInOne, we hold out views 0 and 1 and use the tasks, scored steps, window, and scorer of \cref{tab:x_eval}, with one time step per native frame. The fusion of all views serves as the ground truth, and \ours{} reads the fusion of the observed views. We report no novel-view image metrics on these datasets. PhysGaia and Kubric-4D are licensed under CC BY-NC 4.0, so we do not redistribute converted data.

\FloatBarrier
\begin{table*}[!htbp]
  \caption{\textbf{Rendered predictions, all views and references.} Chamfer distance (mm) and image metrics on observed (Obs.) and novel views of the comparison scenes. Best and second best are marked among the methods only.}
  \label{tab:x_img_full}
  \small%
  \centering%
  \setlength{\tabcolsep}{4pt}%
\begin{tabular*}{\textwidth}{@{\extracolsep{\fill}}l*{7}{c}@{}}
    \toprule
    & 3D & \multicolumn{2}{c}{Silhouette IoU \up} & \multicolumn{2}{c}{PSNR \up} & SSIM \up & LPIPS \down \\
    \cmidrule(lr){2-2}\cmidrule(lr){3-4}\cmidrule(lr){5-6}\cmidrule(lr){7-7}\cmidrule(l){8-8}
    Method & CD \down & Obs. & Novel & Obs. & Novel & Novel & Novel \\
    \midrule
    \multicolumn{8}{@{}l}{\textit{In-between prediction}} \\
    FreeGave~\cite{li2025freegave} & \second{28.7} & \second{0.796} & \second{0.815} & \best{29.01} & \best{21.45} & \second{0.786} & \second{0.249} \\  
    TRACE~\cite{li2025trace} & 96.9 & 0.460 & 0.489 & 24.21 & 19.76 & 0.716 & 0.340 \\  
    \oursrow \ours{} & \best{6.0} & \best{0.819} & \best{0.828} & \second{27.97} & \second{21.22} & \best{0.801} & \best{0.225} \\  
    \cmidrule(l){1-8}
    Voxel ceiling & 0.0 & 0.944 & 0.936 & 33.91 & 22.13 & 0.821 & 0.211 \\  
    \midrule
    \multicolumn{8}{@{}l}{\textit{Future prediction}} \\
    FreeGave~\cite{li2025freegave} & \second{65.7} & \second{0.349} & \second{0.388} & \best{22.37} & \best{20.37} & \second{0.790} & \second{0.250} \\  
    TRACE~\cite{li2025trace} & 131.7 & 0.299 & 0.286 & 21.77 & 19.74 & 0.730 & 0.333 \\  
    \oursrow \ours{} & \best{36.9} & \best{0.504} & \best{0.526} & \second{21.88} & \second{19.96} & \best{0.807} & \best{0.226} \\  
    \cmidrule(l){1-8}
    Voxel ceiling & 0.0 & 0.941 & 0.927 & 27.56 & 22.45 & 0.857 & 0.172 \\  
    \bottomrule
  \end{tabular*}%
\end{table*}

\begin{table*}[!htbp]
  \caption{\textbf{Adapted 4D methods.} F1 (\%, \up) of the adapted 4D methods on the comparison scenes: primary candidates (top), and released settings and variants (bottom).}
  \label{tab:juryrig_main}
  \small%
  \centering%
  \setlength{\tabcolsep}{1pt}%
\begin{tabular*}{\textwidth}{@{\extracolsep{\fill}}ll*{12}{c}@{}}
    \toprule
    & & \multicolumn{7}{c}{In-between prediction} & \multicolumn{5}{c}{Future prediction} \\
    \cmidrule(lr){3-9}\cmidrule(l){10-14}
    Method & Input / variant
      & \shortstack{Elastic\\Fall} & \shortstack{DropIn\\Liquid} & \shortstack{Crank\\Push} & \shortstack{Panel\\Spin} & \shortstack{Double\\Pendulum} & \shortstack{Balloon\\Lift} & Mean
      & \shortstack{Elastic\\Fall} & \shortstack{Panel\\Spin} & \shortstack{Double\\Pendulum} & \shortstack{Balloon\\Lift} & Mean \\
    \midrule
    FreeGave & RGB, D\textsubscript{0}, M
      & 77.1 & 21.7 & \second{85.2} & 56.3 & 91.5 & 82.8 & 69.1 & 0.7 & 18.6 & 33.6 & 73.3 & 31.5 \\
    TRACE & RGB, D\textsubscript{0}, M
      & 0.0 & 15.1 & 70.2 & 32.5 & 15.7 & 45.8 & 29.9 & 2.5 & 32.8 & 34.4 & 44.0 & 28.4 \\
    SoM & RGB-D, M, 1 view
      & 77.5 & 49.1 & 73.5 & 72.2 & 82.2 & 49.3 & 67.3 & 13.5 & 23.8 & 44.0 & 43.5 & 31.2 \\
    MoSca & RGB-D, M, 1 view
      & 73.5 & 19.5 & 40.6 & 33.6 & 52.2 & 40.3 & 43.3 & 5.9 & 31.0 & 27.1 & 50.6 & 28.6 \\
    STv2 & RGB-D, M
      & \second{96.4} & \second{65.1} & 84.7 & \best{89.2} & \best{96.1} & \second{95.5} & \second{87.8} & 4.2 & \best{57.4} & 50.4 & \best{90.8} & \second{50.7} \\
    OpenD4RT & RGB-D, M
      & 93.2 & 61.6 & 82.2 & 77.3 & 94.5 & 95.3 & 84.0 & \second{16.5} & 30.7 & \second{63.4} & 73.1 & 45.9 \\
    \oursrow \ours{} & RGB-D, M
      & \best{97.3} & \best{77.9} & \best{87.2} & \second{86.3} & \second{95.4} & \best{95.8} & \best{90.0} & \best{22.0} & \second{47.6} & \best{80.7} & \second{83.2} & \best{58.4} \\
    \midrule
    \multicolumn{14}{@{}l}{\textit{Released settings and variants}} \\
    SoM & released priors
      & 62.8 & 36.7 & 66.2 & 51.6 & 54.8 & 49.3 & 53.6 & 35.8 & 28.3 & 44.6 & 43.5 & 38.1 \\
    MoSca & released profile
      & 65.9 & 18.7 & 37.7 & 24.7 & 36.8 & 39.3 & 37.2 & 9.4 & 30.6 & 41.3 & 45.0 & 31.6 \\
    STv2 & stride 4 on all
      & 96.2 & 64.5 & 83.4 & 87.4 & 96.0 & 95.5 & 87.2 & 4.6 & 54.8 & 50.1 & 90.6 & 50.0 \\
    OpenD4RT & own geometry
      & 86.7 & 24.2 & 76.0 & 47.8 & 75.9 & 61.7 & 62.1 & 18.8 & 23.2 & 54.9 & 51.2 & 37.0 \\
    \bottomrule
  \end{tabular*}%
\end{table*}
\begin{table*}[!htbp]
  \caption{\textbf{Motion split on PhysInOne.} F1 (\%, \up) on ground-truth-defined target subsets: linear and non-linear in-between, and future by horizon and contact. $n$: scored steps.}
  \label{tab:juryrig_split}
  \centering
  \small
  \setlength{\tabcolsep}{2.0pt}%
\begin{tabular*}{\textwidth}{@{\extracolsep{\fill}}l*{7}{c}@{}}
    \toprule
    & \multicolumn{3}{c}{In-between} & \multicolumn{4}{c}{Future} \\
    \cmidrule(lr){2-4} \cmidrule(l){5-8}
    Method & linear & non-lin. & \shortstack{non-lin.\\w/o PS} & $j{\leq}3$ & $j{\geq}4$ & \shortstack{before\\contact} & \shortstack{after\\contact} \\
    & $n{=}89$ & $n{=}45$ & $n{=}19$ & $n{=}6$ & $n{=}28$ & $n{=}26$ & $n{=}8$ \\
    \midrule
    FreeGave~\cite{li2025freegave} & 78.2 & 46.5 & 33.6 & 62.3 & 25.7 & 41.7 & 1.3 \\
    TRACE~\cite{li2025trace} & 21.3 & 26.3 & 20.3 & 58.2 & 23.5 & 37.1 & 5.5 \\
    SoM~\cite{wang2025shape} & 75.1 & 64.1 & 52.1 & 65.1 & 25.2 & 38.2 & 13.1 \\
    MoSca~\cite{lei2025mosca} & 49.3 & 30.1 & 25.0 & 51.6 & 24.5 & 36.2 & 6.8 \\
    STv2~\cite{xiao2025spatialtrackerv2} & 95.1 & 78.3 & 64.7 & 88.9 & 44.4 & 65.8 & 8.2 \\
    OpenD4RT & 90.3 & 71.0 & 61.9 & 77.4 & 40.9 & 56.8 & 16.6 \\
    \oursrow \ours{} & 94.7 & 80.6 & 75.7 & 85.0 & 55.2 & 71.8 & 23.5 \\
    \midrule
    Oracle linear interpolation (GT states) & 98.2 & 69.9 & 68.6 & 80.1 & 34.0 & 53.3 & 6.0 \\
    \bottomrule
  \end{tabular*}%
\end{table*}
\begin{table*}[!htbp]
  \caption{\textbf{Physics benchmarks, per scene.} Per-scene F1 (\%, \up) on the common subset of \cref{tab:physbench} (top) and on the remaining scenes, without FreeGave and TRACE (bottom). $n$: scored targets. \textsuperscript{\ddag}: excluded, as its F1 on the observed steps is below 90.}
  \label{tab:physbench_base}
  \small%
  \centering%
  \setlength{\tabcolsep}{2.2pt}%
\begin{tabular*}{\textwidth}{@{\extracolsep{\fill}}lrcccccc>{\columncolor{oursbg}}c@{}}
    \toprule
    Scene & $n$ & FreeGave & TRACE & SoM & MoSca & STv2 & OpenD4RT & \ours{} \\
    \midrule
    \multicolumn{9}{@{}l}{\textit{In-between}} \\
    \multicolumn{9}{@{}l}{\quad common subset (\cref{tab:physbench})} \\
    torus & 13 & 52.9 & 85.5 & 79.9 & 47.6 & 93.3 & 88.3 & 92.2 \\
    cross & 12 & 80.0 & 86.4 & 72.3 & 56.3 & 87.7 & 87.8 & 88.5 \\
    apple & 11 & 62.0 & 70.5 & 56.8 & 68.4 & 89.3 & 91.3 & 90.6 \\
    PhysGaia balls & 26 & 47.9 & 48.1 & 50.2 & 27.5 & --\textsuperscript{\ddag} & 74.6 & 78.5 \\
    scn02700 & 13 & 70.5 & 60.4 & 76.7 & 41.6 & 99.0 & 97.9 & 98.4 \\
    scn02800 & 14 & 45.8 & 50.3 & 73.5 & 37.8 & 98.8 & 96.8 & 97.6 \\
    scn02801 & 13 & 36.1 & 41.5 & 69.5 & 29.6 & 99.6 & 97.6 & 99.1 \\
    \multicolumn{9}{@{}l}{\quad remaining scenes} \\
    cream & 3 &  &  & 78.4 & 42.6 & 91.5 & 90.7 & 93.6 \\
    toothpaste & 9 &  &  & 75.5 & 29.8 & 95.2 & 91.5 & 95.0 \\
    chess & 12 &  &  & 75.3 & 57.6 & 87.1 & 85.7 & 86.9 \\
    banana & 12 &  &  & 71.7 & 50.0 & 86.1 & 86.8 & 87.5 \\
    scn02802 & 8 &  &  & 70.2 & 45.5 & 98.9 & 96.7 & 97.5 \\
    scn02803 & 0 &  &  &  &  &  &  &  \\
    scn02804 & 8 &  &  & 58.9 & 33.1 & 98.2 & 94.6 & 98.4 \\
    \midrule
    \multicolumn{9}{@{}l}{\textit{Future}} \\
    \multicolumn{9}{@{}l}{\quad common subset (\cref{tab:physbench})} \\
    torus & 10 & 39.1 & 22.4 & 58.7 & 24.3 & 51.6 & 46.5 & 59.8 \\
    cross & 9 & 61.9 & 54.5 & 63.2 & 36.3 & 54.8 & 59.0 & 63.4 \\
    apple & 8 & 37.2 & 20.4 & 27.6 & 21.4 & 41.8 & 31.5 & 44.3 \\
    PhysGaia balls & 7 & 56.1 & 67.8 & 48.0 & 36.4 & --\textsuperscript{\ddag} & 65.8 & 62.2 \\
    scn02700 & 8 & 71.6 & 62.9 & 64.5 & 49.9 & 89.8 & 72.4 & 83.6 \\
    scn02800 & 8 & 44.8 & 48.5 & 53.2 & 46.7 & 85.7 & 70.4 & 78.0 \\
    scn02801 & 6 & 72.9 & 61.7 & 58.7 & 19.2 & 83.5 & 74.0 & 81.0 \\
    \multicolumn{9}{@{}l}{\quad remaining scenes} \\
    cream & 2 &  &  & 58.8 & 23.0 & 49.9 & 57.3 & 64.6 \\
    toothpaste & 9 &  &  & 57.8 & 40.8 & 61.6 & 63.6 & 62.3 \\
    chess & 8 &  &  & 43.7 & 33.0 & 44.4 & 42.9 & 46.5 \\
    banana & 9 &  &  & 38.9 & 7.5 & 34.2 & 19.4 & 59.3 \\
    scn02802 & 7 &  &  & 54.6 & 55.6 & 85.2 & 72.2 & 77.9 \\
    scn02803 & 1 &  &  & 74.3 & 61.4 & 93.0 & 88.3 & 89.5 \\
    scn02804 & 6 &  &  & 46.3 & 47.3 & 90.0 & 79.0 & 87.4 \\
    \bottomrule
  \end{tabular*}%
\end{table*}
\begin{table*}[!htbp]
  \caption{\textbf{Fusion, network, integration, and rendering.} Settings of \ours.}
  \label{tab:x_impl}
  \centering
  \small
  \begin{tabular}{@{}p{0.2\textwidth}p{0.76\textwidth}@{}}
    \toprule
    Setting & Value \\
    \midrule
    \multicolumn{2}{@{}l}{\emph{Fusion}} \\
    Input & static cameras at 560\,px (RGB box-downsampled from 1120\,px), every second video frame \\
    Moving-object pixels & segmentation label $\geq 128$ or tagged as moving \\
    Cameras & OpenGL convention, $\operatorname{diag}(1,-1,-1)$ before $(\bm{R}^n, \bm{t}^n)$, pixel centers at index $+\tfrac{1}{2}$ \\
    Voxel size $\Delta$ & 1.25\,cm (PhysGaia: scene scaled by 0.2, Kubric-4D: 10\,cm) \\
    Occupancy & at least 3 points per voxel \\
    Color, objectness & normalized trilinear weights at occupied voxels \\
    Unseen flag & unoccupied, and outside the image or more than $\Delta/2$ behind the observed depth in every camera \\
    Grid extent & 0.5--99.5 percentile box of the ground-truth moving-object points of the clip, plus 5\,cm (axes at most 4\,m) \\
    \midrule
    \multicolumn{2}{@{}l}{\emph{Network}} \\
    Architecture & 3D U-Net, widths 32, 64, 96, 128, blocks of ($3^3$ convolution, LayerNorm, SiLU) $\times\,2$ \\
    Input & 21 channels: three fused states ($3 \times 6$), $e$, $s/10$, and a zero time channel, color and objectness divided by 0.359 \\
    Output & 13 node fields $\V_0, \dots, \V_{12}$ ($H = 12$), zero-initialized \\
    Size, precision & 2,879,335 parameters, bfloat16 forward pass, float32 advection and losses \\
    \midrule
    \multicolumn{2}{@{}l}{\emph{Integration}} \\
    Substeps & midpoint scheme with both stages at the substep mid-time, 4 per time step in training and 8 at evaluation \\
    Large windows & $128^3$ tiles at stride 64 (cross-scene: $64^3$ tiles, 12-voxel margins) \\
    \midrule
    \multicolumn{2}{@{}l}{\emph{Rendering}} \\
    Object pixels & rendered as points at 560\,px, depth tolerance 2\%, holes closed if at least 5 of 8 neighbors are covered, opacity from $2 \times 2$ coverage \\
    Static image & training views: nearest observed non-object pixel, novel views: depth points of the last 3 observations, weighted by $\exp(-\theta/15^\circ)$ \\
    \bottomrule
  \end{tabular}
\end{table*}
\begin{table*}[!htbp]
  \caption{\textbf{Training.} Training settings of \ours.}
  \label{tab:appendix_hparams}
  \centering
  \small
  \begin{tabular}{@{}p{0.2\textwidth}p{0.76\textwidth}@{}}
    \toprule
    Setting & Value \\
    \midrule
    \multicolumn{2}{@{}l}{\emph{Data}} \\
    Scenes & 152 PhysInOne training scenes (11,059 steps, tiers A 26, B1 45, B2 18, C 23, D 40), those with fused states at the start of training \\
    Split & 10\% of each tier held out (seed 0), no granular or light phenomena \\
    \midrule
    \multicolumn{2}{@{}l}{\emph{Optimization}} \\
    Steps, batch & 24,000 steps, 4 crops of $48^3$ voxels \\
    Optimizer & Adam~\cite{kingma2015adam}, learning rate $10^{-3}$, 500 warmup steps, cosine decay, gradient clipping 1.0, seed 0 \\
    \midrule
    \multicolumn{2}{@{}l}{\emph{Sampling}} \\
    Spacing $s$ & $\{1,2,3,4,6,8,11\}$, at most a horizon cap that grows from 1 to 12 over 7,200 steps \\
    Horizon $h$ & next observation given with probability 0.75 ($h = s$), otherwise $h = s$ or uniform up to the cap \\
    Evaluated settings & with probability 0.5, $(s, h, e) \in \{(2,2,1), (4,4,1), (8,8,1), (2,10,0)\}$ \\
    Scene, initial observation & with probability 0.5 by squared moving average of the loss, otherwise uniform, initial observation at a multiple of $s$ \\
    Redraw, crop & redrawn until at least 50 object voxels move, crop centered on a moved voxel, jitter $\pm 12$ voxels \\
    \midrule
    \multicolumn{2}{@{}l}{\emph{Losses}} \\
    Points & up to 1,024 source and target points per step \\
    $\ell_{point}$, $\ell_{rigid}$ & Euclidean Chamfer distance in voxels, floor 0.5 voxel, $\lambda = 0.2$ from step 9,600, 6 neighbors \\
    \midrule
    \multicolumn{2}{@{}l}{\emph{Run}} \\
    Hardware & one RTX 5080 (16\,GB), 25\% memory cap \\
    Checkpoint & last (step 24,000), no selection, resumed once from step 12,000 \\
    \bottomrule
  \end{tabular}
\end{table*}
\begin{table*}[!htbp]
  \caption{\textbf{Fused inputs estimated from RGB.} Settings of the input fusions of \cref{sec:x_exp_input}.}
  \label{tab:x_lifts}
  \centering
  \small
  \begin{tabular}{@{}p{0.2\textwidth}p{0.76\textwidth}@{}}
    \toprule
    Setting & Value \\
    \midrule
    Stereo depth & plane sweep~\cite{collins1996space}, 192 then 32 hypotheses in inverse depth, ZNCC ($7 \times 7$) over the best 4 views, kept if the score exceeds 0.35 and the depth agrees within 1\% in at least 2 views \\
    Stereo labels & background subtraction on a 2.5\,cm grid: seeds where occupancy changes at least 3 times, background occupied in at least 70\% of the frames, objects as foreground components touching a seed \\
    Labeling frames & the observations (in-between), the first half of the clip (future), which includes the first predicted step in five scenes \\
    Estimated depth & Depth Anything~3~\cite{lin2026depth} (giant), on all training views of a step with our cameras, metric scale from our cameras \\
    Estimated masks & SAM~2.1~\cite{ravi2025sam}, prompted automatically from color and depth changes, video propagation over the observed steps \\
    \bottomrule
  \end{tabular}
\end{table*}
\begin{table*}[!htbp]
  \caption{\textbf{Evaluation protocol.}}
  \label{tab:x_eval}
  \centering
  \small
  \begin{tabular}{@{}p{0.2\textwidth}p{0.76\textwidth}@{}}
    \toprule
    Setting & Value \\
    \midrule
    Views & views 0 and 1 held out; the others observed (10; 13 for ElasticFall) \\
    Scored steps & the ground-truth object moved by more than one voxel since the previous observation (symmetric mean nearest-neighbor distance); the same steps for all methods \\
    Window & percentile box of the moving voxels of the full-view ground-truth fusion, padded; capped at 128 voxels in the cross-scene evaluation \\
    Ground truth & object voxels of the fusion from all cameras \\
    Predictions & ours: advected points rounded to voxels; baselines: rendered depth back-projected where opacity and objectness exceed 0.5, at least 3 hits per voxel \\
    Metrics & F1 at $\tau = 2$ voxels (25\,mm); Chamfer distance $\frac{1}{2}(\mathrm{med} + \mathrm{med})$ in mm (cross-scene: means); silhouette IoU, PSNR, SSIM, and LPIPS (VGG) at 280\,px \\
    Aggregation & mean over the scored steps of a scene, then over scenes \\
    References & Linear: per-voxel blend of two observations, threshold 0.5; Voxel ceiling: fusion of the target step from the observed views \\
    \bottomrule
  \end{tabular}
\end{table*}
\begin{table*}[!htbp]
  \caption{\textbf{Comparison scenes.} Per-scene protocol: time steps $T$, training views, evaluation window, and scored time steps.}
  \label{tab:appendix_scenes}
  \centering
  \small
  \setlength{\tabcolsep}{4pt}%
\begin{tabular*}{\textwidth}{@{\extracolsep{\fill}}lcccrr@{}}
    \toprule
    Scene & $T$ & Views & Window (voxels) & In-between & Future \\
    \midrule
    ElasticFall    & 45 & 13 & $32\times56\times64$   & 16 & 7 \\
    DropInLiquid   & 75 & 10 & $304\times232\times112$ & 17 & 0 \\
    CrankPush      & 75 & 10 & $176\times48\times56$  & 3  & 0 \\
    PanelSpin      & 75 & 10 & $240\times224\times56$ & 48 & 9 \\
    DoublePendulum & 75 & 10 & $64\times72\times96$   & 41 & 10 \\
    BalloonLift    & 75 & 10 & $96\times72\times120$  & 9  & 8 \\
    \midrule
    Total & & & & 134 & 34 \\
    \bottomrule
  \end{tabular*}%
\end{table*}
\begin{table*}[!htbp]
  \caption{\textbf{Per-scene dynamics baselines.} Settings of FreeGave~\cite{li2025freegave} and TRACE~\cite{li2025trace}.}
  \label{tab:x_baselines}
  \centering
  \small
  \begin{tabular}{@{}p{0.2\textwidth}p{0.76\textwidth}@{}}
    \toprule
    Setting & Value \\
    \midrule
    Code & PhysInOne adaptations, 25,000 iterations. FreeGave: physics code 16. TRACE: \texttt{-{}-is\_blender -{}-light} \\
    Resolution & 280\,px (PhysInOne: 1120\,px) \\
    Inputs & RGB of the observed training views, Gaussians from the ground-truth depth of the first frame, masks label object Gaussians for scoring \\
    Time & future: time $=$ frame$/(N-1)$, trained up to the last observation. In-between: times scaled by 0.75 \\
    Near plane & 0.8, or 0.95 times the nearest training depth if at least 1\% of the pixels are nearer \\
    Other changes & opacity reset disabled where training collapsed, TRACE focal lengths set at inference, activation checkpointing \\
    Not compared & NVFi and VeloGauss (no PhysInOne adaptation), NVD (no code), Uni4R (no code, formulation study in \cref{sec:x_uni4r}), D4RT (no weights, OpenD4RT used), Trace Anything (GPU memory), 3DGSim and PGND \\
    \bottomrule
  \end{tabular}
\end{table*}
\begin{table*}[!htbp]
  \caption{\textbf{Adapting 4D methods.} Settings of the adapted 4D reconstruction and tracking methods, which read only the observed views at observed steps. D: depth, M: moving-object masks, C: cameras.}
  \label{tab:juryrig_took}
  \small
  \centering
  \begin{tabular}{@{}p{0.1\textwidth}p{0.195\textwidth}p{0.195\textwidth}p{0.195\textwidth}p{0.195\textwidth}@{}}
    \toprule
    & SoM~\cite{wang2025shape} & MoSca~\cite{lei2025mosca} & STv2~\cite{xiao2025spatialtrackerv2} & OpenD4RT \\
    \midrule
    Type & per-scene fit & per-scene fit & feed-forward & feed-forward (unofficial weights) \\
    Input & RGB, D, M, C of one view & RGB, D, M, C of one view & RGB, D, C of every view, queries on M & RGB, C of every view, queries on M, D for scale, geometry from our fusion \\
    View & most moving-object pixels & most moving-object pixels & all, points united & all, points united \\
    Time & motion bases blended, constant velocity (future) & nodes: linear positions, slerped rotations, constant velocity (future) & per-track linear, constant velocity (future) & per-track linear, constant velocity (future) \\
    Changes & motion priors off (candidate), mask erosion 3 & velocity regularizers off (candidate), keyframe counts for short clips & ground-truth cameras, 128 queries per pass, query stride 2 (candidate) & observed frames repeated (candidate), 16-frame windows \\
    Points & rendered depth, alpha and objectness above 0.5, 3 hits & as SoM & tracked points, hit count chosen on observed steps & as STv2 \\
    \midrule
    Selection & \multicolumn{4}{p{0.8\textwidth}@{}}{primary candidate chosen by F1 on observed steps only} \\
    \bottomrule
  \end{tabular}
\end{table*}
\begin{table*}[!htbp]
  \caption{\textbf{Physics benchmarks, setup.} All protocol choices, including the contact split, were fixed before any method was run.}
  \label{tab:x_bench}
  \centering
  \small
  \begin{tabular}{@{}p{0.12\textwidth}p{0.26\textwidth}p{0.26\textwidth}p{0.26\textwidth}@{}}
    \toprule
    & Spring-Gaus~\cite{zhong2024reconstruction} & PhysGaia~\cite{kim2026physgaia} & Kubric-4D~\cite{vanhoorick2024gcd,greff2022kubric} \\
    \midrule
    Content & elastic objects dropped onto a plane (MPM) & five viscoelastic balls falling into a bowl & rigid objects tossed onto others \\
    Frames, views & 30 frames, 10 views & frames 26--115, 6 views & 60 frames, 16 views \\
    Depth, masks & rendered from the particles & rendered from the particles and scene & the dataset's \\
    Scale & 1.25\,cm voxels & scene scaled by 0.2 & 10\,cm voxels \\
    Scenes & ours 15, adapted methods 7, FreeGave and TRACE 3 & one, plus two stress scenes & six, FreeGave and TRACE 3 \\
    Contact & lowest particle within 1.2\,cm of the ground & ball acceleration departs from free fall and rest by more than $g/2$ & onset of a collision above the 75th force percentile \\
    \bottomrule
  \end{tabular}
\end{table*}

\section*{Code, Data, and Credits}
\label{sec:supplemental_materials}
Upon publication, we will release the code of \ours{} (fusion, training, inference, rendering, and evaluation), the trained model used for all results in this paper, the scene lists of our training, test, and validation splits, the per-scene results of every method as JSON files, and the scripts and settings with which we fit, adapt, render, and score the baselines.
The PhysInOne~\cite{zhou2026physinone}, Spring-Gaus~\cite{zhong2024reconstruction}, PhysGaia~\cite{kim2026physgaia}, and Kubric-4D~\cite{vanhoorick2024gcd} scenes are available from their authors.
The scene images in \cref{fig:teaser,fig:pipeline,fig:qualitative} and \appref{fig:qual_gallery} and \appref{fig:qual_compare} are renderings of PhysInOne scenes, or are rendered by \ours{} and the compared methods from observations of these scenes.

\end{document}